\documentclass{article}

\usepackage{arxiv}

\usepackage[utf8]{inputenc} % allow utf-8 input
\usepackage[T1]{fontenc}    % use 8-bit T1 fonts
\usepackage{hyperref}       % hyperlinks
\usepackage{url}            % simple URL typesetting
\usepackage{booktabs}       % professional-quality tables
\usepackage{amsfonts}       % blackboard math symbols
\usepackage{nicefrac}       % compact symbols for 1/2, etc.
\usepackage{microtype}      % microtypography
\usepackage{lipsum}
\usepackage{graphicx}
\usepackage{amsmath} 
\usepackage[ruled,vlined]{algorithm2e}
\graphicspath{ {./images/} }

\title{Interpretable Hybrid Deep Q-Learning Framework for IoT-Based Food Spoilage Prediction with Synthetic Data Generation and Hardware Validation}

\author{
 Isshaan Singh \\
  School of Computer Science and Engineering \\
  Vellore Institute of Technology \\
  Chennai, Tamil Nadu 600127, India \\
  \texttt{isshaan.sing2003@gmail.com} \\
 \And
 Divyansh Chawla \\
  School of Computer Science and Engineering \\
  Vellore Institute of Technology \\
  Chennai, Tamil Nadu 600127, India \\
  \texttt{divyanshchawla496@gmail.com} \\
 \And
 Anshu Garg \\
  School of Computer Science and Engineering \\
  Vellore Institute of Technology \\
  Chennai, Tamil Nadu 600127, India \\
  \texttt{aishashu2003@gmail.com} \\
 \And
 Shivin Mangal \\
  School of Computer Science and Engineering \\
  Vellore Institute of Technology \\
  Chennai, Tamil Nadu 600127, India \\
  \texttt{shivin.m17@gmail.com} \\
 \And
 Pallavi Gupta \\
  School of Computer Science and Engineering \\
  Vellore Institute of Technology \\
  Chennai, Tamil Nadu 600127, India \\
  \texttt{Pallavirkg2004@gmail.com} \\
 \And
 Khushi Agarwal \\
  School of Computer Science and Engineering \\
  Vellore Institute of Technology \\
  Chennai, Tamil Nadu 600127, India \\
  \texttt{Khushiagarwal835@gmail.com} \\
 \And
 Nimrat Singh Khalsa \\
  School of Computer Science and Engineering \\
  Vellore Institute of Technology \\
  Chennai, Tamil Nadu 600127, India \\
  \texttt{nimratsingh.khalsa@gmail.com} \\
 \And
 Nandan Patel \\
  School of Computer Science and Engineering \\
  Vellore Institute of Technology \\
  Chennai, Tamil Nadu 600127, India \\
  \texttt{nandanpatel2411@gmail.com} \\
}

\begin{document}
\maketitle
\begin{abstract}
The need for an intelligent, real-time spoilage prediction system has become critical in modern IoT-driven food supply chains, where perishable goods are highly susceptible to environmental conditions. Existing methods often lack adaptability to dynamic conditions and fail to optimize decision-making in real-time. To address these challenges, we propose a hybrid reinforcement learning framework integrating Long Short-Term Memory (LSTM) and Recurrent Neural Networks (RNN) for enhanced spoilage prediction. This hybrid arrangement effectively captures temporal dependencies within sensor data, ensuring robust and adaptive decision-making. In alignment with the principles of interpretable artificial intelligence, a rule-based classifier environment has been employed to provide transparent ground truth labeling for spoilage levels based on interpretable domain-specific thresholds. This structured design enables the agent to operate within clearly defined semantic boundaries, facilitating traceable decision-making and aiding model interpretability. Furthermore, model behaviour is monitored through interpretability-driven metrics such as spoilage accuracy, reward-to-step ratio, loss decrease rate, and exploration rate decay. These metrics offer not only quantitative performance evaluation but also insights into learning dynamics and prediction rationale. A class-wise spoilage distribution chart is utilized to visualize the agent’s decision profile, thereby enabling intuitive assessment of policy behavior. Through extensive evaluations on both simulated and real-time hardware data, our results demonstrate that the LSTM + RNN agent outperforms alternative agentic frameworks in spoilage prediction accuracy and decision efficiency, while maintaining high levels of interpretability. The findings highlight the significance of hybrid deep reinforcement learning with integrated interpretable elements in IoT-based monitoring applications, paving the way for more transparent, adaptive, and scalable solutions in food safety and supply chain management.
\end{abstract}


\section{Introduction}
The increasing demand for efficient real-time monitoring of food transportation systems is driven by the need to reduce food wastage and ensure product quality, as significant spoilage occurs during transit [1]. Traditional monitoring approaches often rely on static methods, which fail to adapt to the dynamic conditions of transportation [2]. Recent advancements in technology, such as the integration of IoT and deep learning, have shown promise in enhancing food quality monitoring by capturing real-time environmental data and predicting spoilage risks [1]. Additionally, innovative solutions like edible electronics and microcontroller-based systems provide accurate, real-time spoilage detection, enabling proactive interventions [3][5]. Furthermore, AI-driven models, such as YOLOv8, facilitate rapid identification of damaged products, thereby improving quality control processes in the food industry [4]. Collectively, these technologies represent a significant step towards more effective and responsive food monitoring systems.
The current landscape of food spoilage monitoring systems reveals significant limitations, primarily due to their reliance on static strategies that delay spoilage detection and compromise product quality. To address these challenges, recent research proposes an integrated system that combines machine learning (ML) and reinforcement learning (RL) with a network of IoT sensors, including MQ3 gas sensors and DHT11 temperature and humidity sensors, for real-time monitoring [1] [6]. This system enhances data preprocessing through techniques like normalization and feature selection, improving model training efficiency [6]. Additionally, innovative approaches such as deep reinforcement learning and ensemble learning have been shown to optimize resource allocation and improve predictive accuracy in dynamic environments [7]. Furthermore, novel sensor systems, such as those utilizing pH and gas detection, facilitate early spoilage detection, promoting food safety and reducing waste [4] [8]. Collectively, these advancements underscore the urgent need for dynamic monitoring capabilities in food quality management.
The contributions of this research extend beyond conventional monitoring approaches. Firstly, it introduces a novel dynamic optimization methodology using RL to adapt the monitoring strategy based on evolving conditions during transportation. Secondly, it explores feature selection and optimization techniques within the realm of data preprocessing, enhancing the accuracy and efficiency of the monitoring system. Thirdly, the research evaluates the system's performance based on some self-defined key metrics. Lastly, the proposed system holds the potential to significantly reduce food wastage, ensuring the timely delivery of fresh produce to consumers. 

\subsection{Motivation}
Food spoilage prediction in IoT-based systems necessitates dynamic and transparent decision-making capabilities that traditional machine learning (ML) algorithms often lack. While ML models are effective on static datasets, they are generally limited by their inability to adapt to environmental fluctuations such as varying temperature, humidity, and sensor noise [1]. Furthermore, the opaque nature of such models poses significant barriers to trust and interpretability in critical applications like food safety. Reinforcement Learning (RL), in contrast, offers a sequential decision-making paradigm that inherently models temporal dependencies. This makes RL more suitable for real-time adaptability in dynamic environments. In this work, spoilage prediction is formulated as a rule-based classification problem to incorporate human-understandable decision boundaries, thereby enhancing the interpretability of agent behavior. These domain-specific thresholds act as semantic anchors guiding the agent's learning, enabling traceable and interpretable policy refinement. To further improve the model’s capacity for sequential reasoning and temporal pattern recognition, a hybrid Deep Q-Learning architecture is employed, combining Long Short-Term Memory (LSTM) networks with Recurrent Neural Networks (RNNs). These architectures not only mitigate vanishing gradient issues but also enhance interpretability by making it possible to trace spoilage decisions back to specific sequences of sensor readings. The LSTM-RNN hybrid facilitates the interpretation of evolving spoilage conditions, enabling the framework to dynamically adapt its policies with observable rationale. Additionally, the agent’s decision-making trajectory is monitored using self-defined, interpretability-driven metrics such as spoilage accuracy, reward-to-step ratio, and exploration rate decay. These metrics serve dual purposes: evaluating performance and providing interpretable insights into the agent’s learning behavior. By integrating rule-based knowledge, sequential modeling, and interpretability-focused evaluations, the proposed framework bridges the gap between adaptive intelligence and transparent decision-making. This positions it as a theoretically sound and practically viable solution for real-world deployment in food spoilage detection systems.
The proposed research introduces an interpretable, adaptive framework for real-time food spoilage prediction in IoT environments, making the following key contributions:

\begin{itemize}
    \item \textbf{Rule-Based interpretable Classifier for Pre-Spoilage States:}  
    A transparent, domain-informed rule-based classification system is developed to model spoilage thresholds. This classifier encodes expert knowledge into interpretable rules, which serve both as supervisory signals and as interpretability scaffolds for the agent. These rules allow domain experts to understand and validate the basis of spoilage-related decisions in real time.

    \item \textbf{Hybrid Deep Q-Network Architecture with LSTM and RNN:}  
    A novel architecture integrating LSTM and RNN within a Deep Q-Network framework is proposed, designed to capture long- and short-term temporal dependencies in IoT sensor data. The use of sequential neural models allows for transparent tracing of prediction outcomes to temporal patterns, thus increasing interpretability while ensuring high Q-value prediction accuracy.

    \item \textbf{Integration of interpretability Metrics for Policy Evaluation:}  
    Custom metrics are designed not only to measure performance (e.g., spoilage accuracy, reward-to-step ratio) but also to interpret model behavior during training. These metrics offer visibility into the agent's learning trajectory and decision logic, facilitating post-hoc interpretability and policy auditing.

    \item \textbf{Comprehensive Validation on Synthetic and Real-Time Hardware Data:}  
    The framework is validated on both simulated and hardware-based datasets. Real-time IoT data from environmental sensors is used alongside synthetic data to evaluate the adaptability and interpretability of the model under diverse spoilage scenarios. Cross-comparative analysis ensures that the framework maintains interpretability irrespective of data source variability.

    \item \textbf{Scalability, Robustness, and Modular interpretability:}  
    The modular nature of the rule-based system and neural architecture allows for easy expansion across various food types, spoilage thresholds, and sensor configurations. This adaptability is complemented by a consistent interpretability layer, ensuring that interpretability scales alongside system complexity.

\end{itemize}

By unifying rule-based knowledge engineering, deep sequential modeling, and interpretable RL paradigms, this research provides a principled and scalable solution for food spoilage detection and prevention. The integration of interpretability at every stage—from environment design to policy interpretation—paves the way for transparent, trustworthy, and adaptive IoT systems.

\subsection{Paper Organization}
The remainder of this paper is structured as follows. Related Works reviews existing literature on IoT-based food spoilage detection, rule-based classifiers, and reinforcement learning models, identifying gaps addressed by our proposed framework. Methodology details the system design, including the IoT sensor architecture, reinforcement learning framework formulation, and the integration of LSTM and RNN in the Deep Q-Network. It also outlines the performance metrics used to evaluate the framework's effectiveness. The Analysis section compares the proposed framework with other rule-based classifiers and machine learning methods through simulations and experiments. Reward plots and key performance metrics demonstrate the framework’s superiority in deriving optimal spoilage prevention policies. In Conclusion and Future Work, we summarize the findings, emphasizing the framework's adaptability, scalability, and robustness, while discussing limitations such as synthetic data assumptions and potential noise impacts. Future work aims to expand the framework to additional IoT domains, incorporate diverse sensor modalities, and explore real-world deployment scenarios. This structure ensures a comprehensive and cohesive exploration of the proposed framework and its implications.

\section{Related Works}
The integration of Internet of Things (IoT) technologies in monitoring and predicting food spoilage has gained significant attention, utilizing various sensors such as temperature, humidity, gas, and pH to ensure food safety and quality. Studies highlight the use of smart sensors and edible electronics that provide real-time data on environmental conditions throughout the food supply chain, enabling proactive interventions to reduce spoilage and waste[1] [2] [9]. For instance, IoT systems employing machine learning algorithms have demonstrated high accuracy in predicting spoilage by continuously monitoring critical variables[10]. However, current IoT-based methods face limitations in handling dynamic and noisy data, which can lead to inaccuracies in predictions and alerts. Factors such as environmental variability and sensor malfunctions can complicate data interpretation, necessitating further advancements in data processing techniques to enhance reliability[11]. Real-time monitoring is crucial in food supply chains for reducing food wastage and ensuring product quality. Various sensors such as temperature, humidity, and gas sensors are used to monitor the condition of food throughout the supply chain [12]. The collected data from these sensors is sent to a cloud platform where it is compared against threshold values to determine if the food is spoiled [13]. If spoilage is detected, an alert mail is generated and sent to the user, providing details of the spoilage determination. Additionally, real-time sensors are needed to continuously report chemical and biochemical data of the manufacturing process in industrial food processes. This enables the development of monitoring and control strategies based on the continuous measurement of small molecules in the food production process. Overall, real-time monitoring plays a vital role in reducing food wastage and ensuring the quality of food in supply chains. Technological advancements in food spoilage detection have made significant progress in recent years. One area of focus is the development of sensors for detecting food and environmental toxins. Nanofiber-based materials have emerged as promising candidates for constructing electrochemical sensors, offering advantages such as a large surface area and enhanced electron transfer kinetics [14]. Another area of innovation is the integration of Internet of Things (IoT) and Machine Learning (ML) technologies in food freshness detection. This system utilizes IoT-enabled sensors and ML algorithms to monitor and predict changes in the freshness of food products, enabling timely alerts and necessary actions [15]. Additionally, ratiometric sensors have been developed for the detection of mycotoxins in intricate food matrices, providing accurate quantitative analysis and improving food safety [16]. Furthermore, the development of aggregation-induced emission enhancement (AIEE) active and NIR emissive polymers has enabled the detection of H2S gas produced from food spoilage, offering a promising approach for monitoring H2S in practical food samples [17]. These technological advancements in sensors, IoT, and data analytics have significantly contributed to the field of food spoilage detection, improving food safety and consumer health.
Machine learning algorithms have been integrated into various aspects of food quality assurance, including classification, anomaly detection, and predictive modeling. These applications have shown promising results in enhancing food safety and quality. For example, electronic nose technology combined with machine learning algorithms has been used to accurately identify and assess food based on different odors, overcoming traditional limitations related to subjectivity and time-consuming analysis procedures [18]. Machine learning has also been utilized to optimize food processing techniques, improve production efficiency, reduce waste, and create personalized customer experiences in the food industry [19]. Deep learning, a subset of machine learning, has been applied to address food-related challenges such as food recognition, calorie computation, safety detection, and food supply chain management [20]. Additionally, machine learning methods have been used to predict and estimate the nutrient quality of plant-based foods based on their Nutri-score and micronutrient composition [21]. These studies highlight the potential of machine learning in enhancing food quality assurance across various domains in the food industry. 
Reinforcement Learning (RL) has emerged as a pivotal approach in Internet of Things (IoT) systems, particularly in environments necessitating sequential decision-making. Unlike static machine learning models, RL frameworks dynamically adapt their policies in real-time based on environmental feedback, enabling them to handle complex, nonlinear dynamics effectively[22] [26]. For instance, in energy management, RL has been successfully employed to optimize resource allocation in IoT networks, addressing challenges such as signal distortions and adjacent channel interference through deep reinforcement learning techniques[23]. Additionally, RL has shown promise in predictive maintenance, where it autonomously learns optimal maintenance schedules by interacting with sensor data, thus enhancing decision-making speed and accuracy[24]. These applications underscore RL's potential for food spoilage detection, where real-time adaptations can significantly improve monitoring and response strategies, ultimately leading to better food safety outcomes[25] [26]. 
Reinforcement learning (RL) has been explored in the context of dynamic optimization scenarios, particularly in adapting monitoring strategies during transportation. RL has been used to find the best road network configurations in real-time, reducing average travel time and improving traffic flow [27]. RL has also been applied to continuous time and space settings, using stochastic differential equations to drive the underlying dynamics. Occupation time has been developed as a notion for discounted objectives, and performance-difference and local-approximation formulas have been derived [28]. RL has been proposed as an alternative to other optimization methods, such as mix-integer optimization and evolutionary algorithms, for solving sequential decision problems. It has been demonstrated to effectively optimize radar waveform parameters while considering different processing algorithms [29].
Sequential models like Long Short-Term Memory (LSTM) networks and Recurrent Neural Networks (RNNs) are pivotal in capturing temporal dependencies in sensor data, particularly due to their ability to manage long-term dependencies and mitigate issues such as noise. Traditional Artificial Neural Networks (ANNs) often struggle with sequential data because they lack mechanisms to retain information over extended periods, leading to difficulties in modeling temporal patterns effectively[30]. LSTMs, with their memory cells and gating mechanisms, excel in this regard, allowing them to learn from long sequences without succumbing to vanishing or exploding gradients[30]. For instance, in industrial applications, models like the Dual Temporal Attention Mechanism-based Convolutional LSTM have demonstrated significant improvements in prediction accuracy by effectively capturing dynamic features from long sequences[31]. Additionally, LSTMs have outperformed other models in energy consumption predictions within wireless sensor networks, showcasing their robustness against noise and their superior performance in time-series data analysis[32]. Overall, the advantages of LSTMs and RNNs over traditional ANNs are evident in their enhanced capability to handle complex, noisy, and long-term sequential data[33] [34]. Machine Learning (ML) and Reinforcement Learning (RL) frameworks exhibit distinct strengths and weaknesses in predictive and adaptive decision-making, particularly in dynamic environments like the Internet of Things (IoT). While traditional ML approaches, such as supervised and unsupervised learning, rely heavily on pre-existing training data, they often struggle with the uncertainties and rapid changes characteristic of IoT networks[37]. In contrast, RL frameworks excel in these contexts due to their ability to learn from interactions with the environment, adapting policies in real-time without the need for extensive prior data[36][38]. This adaptability is crucial for rule-based classifiers in IoT, where conditions can fluctuate unpredictably, necessitating continuous policy refinement to optimize performance[35]. Furthermore, RL's capacity for multi-objective optimization allows it to effectively manage conflicting goals in complex IoT scenarios, making it a superior choice for dynamic decision-making compared to traditional ML methods[37].
Studies involving IoT hardware experiments for spoilage detection highlight significant challenges such as real-time data acquisition, sensor noise, and environmental variability. Real-time data collection is crucial for timely spoilage detection, yet it is often hindered by sensor inaccuracies and external factors like temperature fluctuations, which can affect sensor readings and lead to false positives or negatives[1] [6]. Additionally, environmental variability complicates the interpretation of data, as different conditions can alter spoilage indicators, necessitating robust algorithms to filter out noise[39] [41]. Real-world validation of theoretical models, such as those integrating IoT and AI, enhances their reliability by demonstrating effectiveness in diverse conditions, as seen in case studies involving various food types like beef and salmon[40] [41]. This empirical evidence supports the applicability of proposed frameworks, ensuring they are not only theoretically sound but also practically viable in reducing spoilage and waste in the food supply chain[6] [40].

\subsection{Problem Statement}
Food spoilage is a dynamic and complex phenomenon influenced by multiple interdependent factors, including environmental conditions, biochemical reactions, and microbial activity. The perishability of food products, along with the variability of spoilage indicators such as temperature, humidity, gas emissions, and pH levels, makes accurate prediction and monitoring a formidable challenge. These indicators are inherently sensitive to external conditions—ranging from transportation fluctuations and inconsistent storage protocols to sensor drift and noise—rendering traditional machine learning (ML) models insufficient for real-world deployment. Static ML models are typically trained on fixed datasets with assumed distributions and fail to capture the variability inherent in real-time IoT sensor data, often resulting in degraded performance due to unmodeled external perturbations [1][11][39]. For example, environmental noise, missing sensor data, or unforeseen temperature spikes during transit can lead to erroneous predictions, such as false spoilage alerts or missed spoilage detections. These shortcomings necessitate the development of dynamic, robust learning frameworks capable of adapting to non-stationary environments and delivering interpretable decisions under uncertainty. Reinforcement Learning (RL) presents a compelling alternative, offering an agent-based approach that iteratively learns optimal policies through real-time interaction with the environment. Deep Q-Networks (DQNs), a prominent RL variant, are especially suited to sequential decision-making tasks like food spoilage detection, where time-dependent fluctuations in sensor readings must be modeled explicitly. DQNs continuously refine their policies in response to observed outcomes, enabling resilience to variability and enhancing predictive robustness [22][25][36]. The inclusion of interpretability into this framework ensures that each decision made by the agent, such as whether to flag a spoilage event, is traceable to specific state-action pairs and environmental features, thereby improving trust and accountability in high-stakes food safety applications. Furthermore, augmenting the RL setup with sequential models such as Long Short-Term Memory (LSTM) networks and Recurrent Neural Networks (RNNs) significantly enhances the ability to model temporal dependencies and noisy input streams. LSTMs, with their specialized memory gating mechanisms, are particularly effective in capturing extended sequences of sensory patterns such as gradual gas emission build-up or humidity fluctuations, which are critical for early spoilage detection [30][31]. These temporal dynamics, when integrated into a hybrid DQN-LSTM-RNN framework, yield a robust, interpretable model capable of identifying not just that spoilage will occur, but why and when, based on sequential environmental patterns. Such traceability is essential for transparency, allowing stakeholders to validate and interpret the model’s outputs across diverse food supply chain contexts. For instance, in refrigerated logistics, the system can dynamically adjust cooling protocols based on detected spoilage trends while providing interpretable rationales for each control action taken. This interpretable hybrid architecture thus empowers IoT systems to transition from reactive to proactive food quality management, enhancing safety, minimizing waste, and promoting sustainable practices. Overall, by embedding interpretable reinforcement learning and temporal modeling into spoilage detection, the proposed solution addresses the critical limitations of static ML models and paves the way for intelligent, adaptive, and transparent decision-making systems in food supply chain monitoring [25][30][36][39].

\section{Methodology}
The methodology proposed in this research focuses on the development of a Deep Reinforcement Learning (RL) framework for precise food spoilage tracking control in IoT-based environments. The approach begins with the generation of synthetic datasets that emulate real-time conditions by leveraging sensor hardware readings collected under controlled setups. These datasets are designed to be versatile and suitable for diverse spoilage tracking scenarios. The hardware configuration employs an Arduino microcontroller as the core processing unit, interfacing with multiple sensors essential for environmental monitoring. Specifically, the MQ3 sensor is used to detect alcohol emissions, the MQ4 sensor is employed to monitor methane levels indicative of decomposition, while the DHT11 sensor measures ambient temperature and humidity—two critical parameters in spoilage prediction. Additionally, a soil moisture sensor assesses residual moisture content, and a servo motor, governed by the Arduino, acts as a control actuator to initiate corrective actions such as regulating water or refrigeration based on sensor input and learned thresholds. Following the data acquisition phase, a custom environment class is designed to emulate the spoilage tracking process within a simulated reinforcement learning setup. This environment synthesizes sensor inputs and simulates spoilage progression, enabling RL agents to interact, learn, and adapt their control strategies accordingly. Within this environment, agents are trained to make optimal decisions on spoilage prevention actions by observing the evolving state of the environment. A central innovation of the proposed framework lies in the integration of deep sequential learning architectures—namely Long Short-Term Memory (LSTM) networks and Recurrent Neural Networks (RNNs)—within the Deep Q-Network (DQN) structure for Q-value estimation. The LSTM layer encodes long-range temporal dependencies from historical sensor data, effectively capturing slow-developing spoilage indicators. Subsequently, an RNN layer refines these sequence embeddings through recurrent connections that highlight dynamic interdependencies in the signal patterns. This dual-layer sequential modeling architecture significantly enhances the model’s capacity to extract actionable insights from temporal data, supporting high-fidelity decision-making in real-time monitoring systems. The training process employs the Deep Q-Learning algorithm, wherein the agent iteratively updates its Q-values to converge towards an optimal policy that minimizes spoilage risk. The decision-making framework consists of four discrete actions: continuing routine storage protocols (Action 0), initiating a low-level alert for minor monitoring (Action 1), adjusting storage parameters such as temperature or humidity (Action 2), and issuing an emergency alert for immediate intervention (Action 3). Each action is selected based on the current state of the environment, as encoded by sensor readings, and refined through the policy learned over repeated episodes. This methodology enables the system to dynamically adapt its spoilage mitigation strategies, ensuring timely interventions while minimizing unnecessary operational overhead. Furthermore, the interpretability of the model’s action choices is preserved by leveraging interpretable deep learning mechanisms embedded within the LSTM-RNN architecture, allowing system operators to trace each decision back to specific sensor input patterns. The comprehensive workflow of the framework, including hardware integration, environment simulation, learning architecture, and action mapping, is illustrated in Figure 1.

\begin{figure}[htbp]
    \centering
    \includegraphics[width=0.8\linewidth]{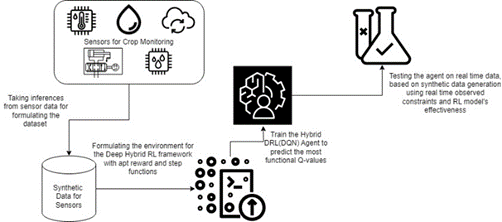} % Change filename and width accordingly
    \caption{ Architecture Diagram of the whole methodology}
    \label{fig:your_label}
\end{figure}

\subsection{Sensor Integration and Synthetic Data Generation}
In the pursuit of precision agriculture and spoilage management, the seamless integration of sensor technologies coupled with innovative synthetic data generation based on real time data monitoring could stand as the cornerstone for revolutionizing food spoilage tracking practices. Figure 2 depicts our data collection setup, showcasing the utilization of various sensors to generate synthetic data. Environmental monitoring, facilitated by sensors like the DHT11 Temperature and Humidity Sensor, empowers informed decision-making by providing real-time insights into critical parameters influencing soil moisture dynamics and plant water usage. Calibration protocols ensure the accuracy of sensor readings, laying the groundwork for reliable data integration. Moreover, the incorporation of gas sensors, such as the MQ3 and MQ4, not only enhances soil health monitoring but also contributes to the preservation of crop freshness by identifying potential issues early on. Leveraging these sensor outputs alongside a meticulously crafted synthetic dataset, the food spoilage tracking system navigates the complex terrain of environmental variability with finesse, paving the way for optimal freshness and yield.  Table 1 provides a description of each sensor employed in our food spoilage tracking optimization framework.

\begin{table}[h]
    \centering
    \caption{Sensors and their details}
    \begin{tabular}{@{}p{5cm}p{10cm}@{}}
        \toprule
        \textbf{Sensor Name} & \textbf{Functionality} \\
        \midrule
        \textbf{DHT11 Temperature and Humidity Sensor} & Continuously measures ambient temperature and humidity, influencing soil moisture dynamics. Higher temperatures increase transpiration, leading to faster moisture depletion, whereas higher humidity reduces transpiration, stabilizing soil moisture. \\[0.3cm]
        \textbf{MQ3 Gas Sensor} & Detects alcohol and other combustible gases, indicating excessive fermentation or decomposition. These conditions negatively impact soil health, affecting nutrient availability and potentially leading to reduced plant growth and compromised freshness. \\[0.3cm]
        \textbf{MQ4 Gas Sensor} & Detects methane gas, providing insights into anaerobic conditions or soil bacteria presence. Methane levels indicate potential soil decomposition under low oxygen conditions, impacting plant respiration and nutrient uptake. \\[0.3cm]
        \textbf{Soil Moisture Sensor} & Directly measures soil water content, providing crucial data for determining food spoilage tracking necessity. Ensures optimal moisture conditions for plant growth, reducing unnecessary watering. \\[0.3cm]
        \textbf{Servo Motor} & Controls the opening and closing of a valve for water release based on sensor data. The system processes soil moisture, temperature, humidity, and gas sensor readings to activate spoilage tracking and optimize moisture conditions. \\
        \bottomrule
    \end{tabular}
    \label{tab:sensors}
\end{table}

\begin{algorithm}[htbp]
\caption{Sensors Setup for Real-Time Data}
\label{alg:sensor_setup}
\SetAlgoLined

\KwIn{DHT.h, Servo.h}
\KwOut{Real-time readings and actuator control}

\textbf{Include Libraries:} DHT.h, Servo.h\;
\textbf{Define Pins:} DHTPIN, MQ3PIN, MQ4PIN, LED1PIN, LED2PIN, LED3PIN, SERVOPIN\;
\textbf{Set Thresholds:} TEMP\_THRESHOLD, MQ3\_THRESHOLD, MQ4\_THRESHOLD\;

\textbf{Initialize:} DHT Sensor and Servo Motor\;

\textbf{Setup:}\;
\Indp
    Initialize Serial Communication\;
    Begin DHT Sensor\;
    Attach Servo to SERVOPIN\;
    Set LED pins as OUTPUT\;
\Indm

\textbf{Loop:}\;
\Indp
    Turn off all LEDs\;
    Read temperature and humidity from DHT sensor\;
    \If{temperature or humidity reading is invalid}{
        Print error message and exit loop\;
    }
    Read analog values from MQ3 and MQ4 sensors\;

    \If{temperature $>$ TEMP\_THRESHOLD}{
        Activate servo motor to 180 degrees\;
        Turn on LED1\;
    }
    \If{MQ3 value $>$ MQ3\_THRESHOLD}{
        Activate servo motor to 90 degrees\;
        Turn on LED2\;
    }
    \If{MQ4 value $>$ MQ4\_THRESHOLD}{
        Activate servo motor to 90 degrees\;
        Turn on LED3\;
    }

    Print temperature, humidity, MQ3, and MQ4 readings\;
    Delay for 2 seconds\;
\Indm
\end{algorithm}

\begin{figure}[htbp]
    \centering
    \includegraphics[width=0.8\linewidth]{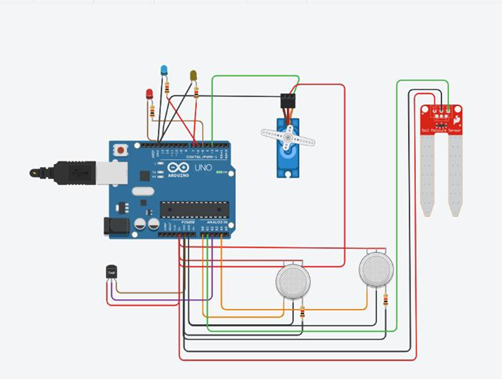} % Change filename and width accordingly
    \caption{Sensors wiring diagram made in Tinkercad}
    \label{fig:your_label}
\end{figure}

Leveraging insights from sensor readings obtained through environmental monitoring, a comprehensive synthetic dataset is meticulously crafted to encompass the diverse range of conditions encountered in agricultural settings, effortlessly adaptable to various use case scenarios. Synthetic data generation initiates by capturing the variability observed in sensor readings—temperature, humidity, and gas concentrations—within specified ranges derived from hardware observations, thereby mirroring the dynamic environmental conditions. To emulate real-world complexities, instances of spoilage thresholds are seamlessly integrated into the dataset, signifying scenarios where monitored parameters exceed predefined thresholds, indicative of potential spoilage or soil health deterioration. This incorporation facilitates capturing the nuanced interplay between environmental factors and crop quality, enabling proactive responses to adverse conditions. Moreover, to emulate the inherent variability and uncertainty observed in real-time sensor data, noise is introduced into the synthetic dataset, modelled using a normal distribution, thereby adding subtle fluctuations mirroring agricultural environments' intrinsic variability. Finally, the synthetic dataset is enriched with food spoilage tracking level labels, meticulously determined based on sophisticated rules derived from real-time hardware monitoring conditions. By analysing sensor readings and predefined spoilage thresholds, the food spoilage tracking system strategically assigns optimal food spoilage tracking strategies tailored to specific environmental conditions, The RL framework is trained to choose the best course of action, which could include continuing normal storage practices (Action 0), issuing a mild alert for monitoring (Action 1), taking proactive steps such as adjusting storage conditions (Action 2), or signaling an urgent need for immediate intervention (Action 3). In conclusion, leveraging real-time monitoring systems, our development of synthetic data paves the way for future IoT-based systems, enabling a more dynamic and responsive approach to agricultural management. Algorithm 1 shows the setup for a generic setting of the sensors in the precise food spoilage tracking setup, which we took for the real time data observed, using temperature-humidity sensor, MQ3 and MQ4 sensor.

\subsection{Deep Hybrid RL Framework formulation}

The Deep Hybrid RL Framework is designed to revolutionize precision agriculture by seamlessly integrating simplicity and scalability into the formulation of environmental parameters. This innovative framework empowers users to easily configure the environment according to specific factors relevant to food spoilage tracking contexts, fostering adaptability and ease of use. Once the environment is prepared, the framework employs a Deep Q-Network (DQN) agent for training purposes. Within the DQN architecture, a sophisticated deep hybrid architecture is utilized. This architecture, comprising a sequential combination of Long Short-Term Memory (LSTM) and Recurrent Neural Network (RNN) layers, exhibits remarkable capabilities in predicting Q-values necessary for agent training. The LSTM component enables the model to capture long-term dependencies in sequential data, while the RNN layer enhances its ability to learn temporal patterns efficiently. The inherent nature of the deep hybrid architecture lends itself to robust performance and a favourable learning curve. By leveraging the strengths of LSTM and RNN layers, the framework facilitates accelerated learning and adaptive decision-making, ultimately enhancing the overall effectiveness of precision food spoilage tracking systems. Figure 3 illustrates the flow of our novel Hybrid DQN model, highlighting its effectiveness in decision-making.
\subsection{Food spoilage tracking Environment Setup}
The state representation of the environment is crucial for precise food spoilage tracking control. At each time step \( t \), the state \( S_t \) is represented as a vector of the following environmental variables:
\begin{figure}[htbp]
    \centering
    \includegraphics[width=0.8\linewidth]{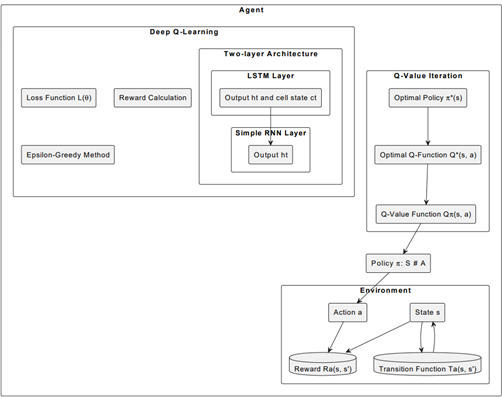} % Change filename and width accordingly
    \caption{Two-Layer Deep Learning Architecture of the DQN Agent for Effective Q-Value Prediction}
    \label{fig:your_label}
\end{figure}

\begin{equation}
    S_t = [\text{Temp}_t, \text{Hum}_t, \text{Moisture}_t, \text{Q3}_t, \text{Q4}_t]
\end{equation}

where:
- \( \text{Temp}_t \) is the temperature at time \( t \),
- \( \text{Hum}_t \) is the humidity at time \( t \),
- \( \text{Moisture}_t \) is the soil moisture at time \( t \),
- \( \text{Q3}_t \) is the MQ3 gas sensor reading at time \( t \),
- \( \text{Q4}_t \) is the MQ4 gas sensor reading at time \( t \).

The synthetic dataset used to simulate real-world sensor readings and environmental conditions is generated by sampling each environmental variable from a normal distribution with specific mean \( \mu_i \) and standard deviation \( \sigma_i \):

\begin{equation}
    X_i \sim \mathcal{N}(\mu_i, \sigma_i^2)
\end{equation}

where \( X_i \) represents the \( i \)-th environmental variable, and \( \mu_i \) and \( \sigma_i \) denote its mean and standard deviation, respectively.
The observation space represents the range of possible states that the RL agent can observe. It is defined as a continuous space bounded between predefined minimum and maximum values for each environmental variable. 

\begin{equation}
    \text{Observation Space} = [0, 1]^5 \tag{3}
\end{equation}

where this notation specifies the bounds of the observation space for each environmental variable. Here, each environmental variable is normalized to a range between 0 and 1.
The exponent 5 indicates that there are five environmental variables in total. The action space for this environment consists of discrete actions corresponding to different food spoilage tracking levels: no food spoilage tracking as 0, low food spoilage tracking as 1, and moderate food spoilage tracking as 2.  The reward function is defined to incentivize the agent to take actions that align with optimal food spoilage tracking strategies. A reward of +1 is assigned if the agent's chosen action matches the true food spoilage tracking level determined by predefined rules. Conversely, a reward of -1 is assigned for mismatching actions. The food spoilage tracking level is determined through predefined rules that leverage sensor readings and spoilage thresholds. These rules are established based on common intuition, although they remain adaptable to different scenarios and needs. By employing condition-based coding or alternative algorithmic approaches, adjustments can be easily made to accommodate specific requirements. This flexibility allows for swift modifications, ensuring the food spoilage tracking strategy remains aligned with evolving environmental conditions and agricultural objectives. For our case, if the temperature exceeds the spoilage threshold and humidity falls below it, moderate food spoilage tracking is recommended. If any sensor reading surpasses its respective spoilage threshold, no food spoilage tracking is advised. Otherwise, low food spoilage tracking is recommended. Through this methodological approach, the environmental setup for the RL framework provides a solid yet simple foundation for training the agent and optimizing food spoilage tracking strategies for enhanced agricultural productivity. Algorithm 2 illustrates the setup and functions of our environment.

\begin{algorithm}[htbp]
\caption{Environment Setup for the Precision Food Spoilage Tracking Context}
\SetKwFunction{ENVSTEP}{ENV\_STEP}
\SetKwFunction{NEXTOBS}{next\_observation}
\SetKwFunction{RESET}{reset}
\SetKwFunction{GETLEVEL}{get\_food\_spoilage\_tracking\_level}
\SetKwFunction{REWARD}{compute\_reward}
\SetKwFunction{DONE}{check\_done}

\SetKwProg{Fn}{Function}{:}{}
\Fn{\ENVSTEP{state, action}}{
    state $\gets$ \NEXTOBS{}\;
    true\_food\_spoilage\_tracking\_level $\gets$ \GETLEVEL{state}\;
    reward $\gets$ \REWARD{true\_food\_spoilage\_tracking\_level, action}\;
    done $\gets$ \DONE{}\;
    \Return{state, reward, done, \{\}}\;
}

\Fn{\NEXTOBS{}}{
    \Return{array of sensor readings for current step}\;
}

\Fn{\RESET{}}{
    Reset current step to 0\;
    \Return{\NEXTOBS{}}\;
}

\Fn{\GETLEVEL{state}}{
    temperature $\gets$ state[0][0]\;
    humidity $\gets$ state[0][1]\;

    \uIf{temperature $>$ spoilage\_thresholds['Temperature'] \textbf{and} humidity $<$ spoilage\_thresholds['Humidity']}{
        \Return{2 \tcp*{Moderate Food Spoilage Tracking}}
    }
    \uElseIf{\texttt{any}(state[0][i][0] $>$ spoilage\_thresholds[key] \textbf{for} i, key \textbf{in} enumerate(spoilage\_thresholds))}{
        \Return{0 \tcp*{No Food Spoilage Tracking}}
    }
    \uElseIf{temperature $>$ spoilage\_thresholds['Temperature'] \textbf{and} humidity $>$ spoilage\_thresholds['Humidity']}{
        \Return{3 \tcp*{High Food Spoilage Tracking - Emergency Action}}
    }
    \Else{
        \Return{1 \tcp*{Low Food Spoilage Tracking}}
    }
}

\Fn{\REWARD{true\_food\_spoilage\_tracking\_level, action}}{
    \eIf{true\_food\_spoilage\_tracking\_level == action}{
        \Return{1}
    }{
        \Return{-1}
    }
}

\Fn{\DONE{}}{
    \eIf{current\_step $\geq$ num\_entries}{
        \Return{True}
    }{
        \Return{False}
    }
}
\end{algorithm}

\subsection{Deep Hybrid Q Learning Network Setup}
The DQN methodology operates within an environment specifically designed for managing food spoilage tracking decisions in IoT-based systems. Here, the agent acts as the decision-maker, interacting with the environment at each time step t. At any given time step t, the agent selects an action a from a defined set of possible actions A. These actions correspond to various food spoilage tracking levels or potential strategies that can be implemented based on the current state s of the environment, encompassing factors like temperature, humidity, soil moisture, and gas sensor readings.
Subsequently, the environment responds by transitioning to a new state s' and providing the agent with a corresponding reward Ra(s, s') . This reward quantifies the effectiveness of the selected action in achieving desired food spoilage tracking outcomes, such as maintaining optimal soil moisture levels for crop growth. The transition to the new state s'  is governed by the state transition function Ta(s, s'), which encapsulates the probability of moving from the current state s to the next state s’ as a result of the agent's chosen action a. This transition probability is influenced by environmental dynamics and the food spoilage tracking action taken, reflecting the complex interplay between environmental factors and food spoilage tracking decisions. 

This decision-making process can be modelled as a Markov Decision Process (MDP), offering a formal framework to understand the interactions between the agent and the environment. 
The MDP model would comprise a tuple (S, A, T, R), where:
- S represents the set of all possible environmental states, encompassing variables like temperature, humidity, soil moisture, and gas sensor readings.
- A denotes the set of all feasible actions the agent can undertake, such as adjusting food spoilage tracking levels or strategies.
In this work, the food spoilage tracking system is formulated as a Markov Decision Process (MDP) where the agent interacts with the environment to optimize food spoilage tracking strategies. The state representation at any time $t$ is described by a vector $S_t$ consisting of various environmental variables, such as temperature, humidity, soil moisture, and gas sensor readings. Specifically, the state at time $t$ is represented as:

\[
S_t = [\text{Temp}_t, \text{Hum}_t, \text{Moisture}_t, \text{Q3}_t, \text{Q4}_t] \quad \text{(1)}
\]

The reward function $R: S \times A \times S \to \mathbb{R}$ provides immediate feedback from the environment, evaluating the action $a$ taken by the agent in transitioning from state $s$ to state $s'$. This feedback quantifies the effectiveness or desirability of the action in achieving food spoilage tracking objectives:

\[
R(s, a, s') \quad \text{(Immediate reward for action $a$ taken in state $s$ resulting in state $s'$)}
\]

The transition probability function $T: S \times A \times S \to [0,1]$ represents the probability of transitioning from state $s$ to state $s'$ when action $a$ is taken. The probability function captures the stochastic nature of environmental dynamics and the uncertainty in state transitions:

\[
T(s, a, s') \quad \text{(Transition probability of moving from state $s$ to state $s'$ after action $a$)}
\]

The agent follows a policy, $\pi: S \to A$, which guides the decision-making process. This policy dictates the agent's behavior based on environmental observations. The goal of reinforcement learning is to discover an optimal strategy that maximizes the cumulative sum of long-term rewards. The cumulative reward over time $t$ is given by:

\[
G_t = R_t + \gamma R_{t+1} + \gamma^2 R_{t+2} + \gamma^3 R_{t+3} + \dots + \gamma^{T-1} R_T = R_t + \sum_{n=1}^T \gamma^n G_{t+1} \quad \text{(4)}
\]

where $R_T$ is the reward received at the terminal state, and $\gamma$ is the discount factor ($0 \leq \gamma \leq 1$), which accounts for the uncertainty in future rewards.

The action-value function, or Q-function, evaluates the value of a state-action pair. For a given state $s$, it determines the desirability of taking action $a$. The Q-function associated with a policy $\pi$ is defined as:

\[
Q_\pi(s, a) = \mathbb{E}_\pi [G_t | S_t = s, A_t = a] \quad \forall s \in S, \forall a \in A \quad \text{(5)}
\]

To find the optimal policy $\pi^*$, the Q-value iteration algorithm is applied. Beginning with an arbitrary state $s$ and action $a$, the algorithm iteratively applies the Bellman equation until the Q-values converge to $Q^*(s, a)$.

The Q-value iteration process is as follows:

\[
Q^{(i)}(s, a) = \sum_{s'} T(s' | s, a) \left[ R(s, a, s') + \gamma \max_{a'} Q^{(i-1)}(s', a') \right] \quad \lim_{i \to \infty} Q^{(i)}(s, a) = Q^*(s, a) \quad \text{(6)}
\]

Once the Q-values converge, the optimal policy $\pi^*$ is determined by:

\[
\pi^*(s) = \arg\max_a Q^*(s, a) \quad \text{(7)}
\]

When the transition probability function or the reward in a Markov decision process is unknown, it becomes a reinforcement learning problem. In Deep Q-learning, deep learning techniques are used to estimate the Q-function. The model receives the current environmental state and generates Q-values for each potential action. The loss function used to predict the Q-function is:

\[
L(\theta) = \left( r(s_t, a_t, s_{t+1}) + \gamma \max_{a_{t+1}} Q(s_{t+1}, a_{t+1}; \theta) - Q(s_t, a_t; \theta) \right)^2 \quad \text{(8)}
\]

This loss function helps the model to optimize the Q-function through training and improve the food spoilage tracking performance.

To estimate the Q-values, a two-layer architecture consisting of LSTM and RNN models with 64 nodes each was employed. The LSTM layer was utilized to capture long-term dependencies in the sequential sensor data, while the subsequent RNN layer refined these representations with recurrent connections. The LSTM layer computes the output $h_t$ and cell state $c_t$ at time step $t$ based on the input $x_t$ and previous state $h_{t-1}$ and $c_{t-1}$.

\[
i_t = \sigma(W_{xi} x_t + W_{hi} h_{t-1} + W_{ci} c_{t-1} + b_i) \quad \text{(9)}
\]
\[
f_t = \sigma(W_{xf} x_t + W_{hf} h_{t-1} + W_{cf} c_{t-1} + b_f) \quad \text{(10)}
\]
\[
g_t = \tanh(W_{xg} x_t + W_{hg} h_{t-1} + W_{cg} c_{t-1} + b_g) \quad \text{(11)}
\]
\[
o_t = \sigma(W_{xo} x_t + W_{ho} h_{t-1} + W_{co} c_{t-1} + b_o) \quad \text{(12)}
\]
\[
c_t = f_t \odot c_{t-1} + i_t \odot g_t \quad \text{(13)}
\]
\[
h_t = o_t \odot \tanh(c_t) \quad \text{(14)}
\]

where $i_t$, $f_t$, and $o_t$ are the input, forget, and output gates, respectively, $g_t$ is the cell input, $\sigma$ is the sigmoid activation function, $\odot$ denotes element-wise multiplication, $W$ are the weight matrices, and $b$ are the bias vectors.

The simple RNN layer computes the output $h_t$ at time step $t$ based on the input $x_t$ and previous state $h_{t-1}$:

\[
h_t = \text{activation}(W_{xh} x_t + W_{hh} h_{t-1} + b_h) \quad \text{(15)}
\]

where $W_{xh} x_t$ and $W_{hh} h_{t-1}$ are the weight matrices for the input and recurrent connection, respectively, $b_h$ is the bias vector, and the activation function is often ReLU or tanh.

These formulas represent the forward pass computations of the LSTM and Simple RNN layers, which process sequential input data and capture temporal dependencies in the DQN agent's neural network model. This combination allowed the model to effectively learn intricate temporal patterns within the environmental data. The model receives the current state of the environment and outputs the Q-value corresponding to each possible action.

Exploration versus exploitation is a crucial consideration in Reinforcement Learning. To balance exploration and exploitation, the Epsilon-Greedy method was employed. Specifically, the agent chooses a random action with probability $\epsilon$ and takes the optimal action with probability $1 - \epsilon$. The $\epsilon$ value was initialized to 1 and decreased by a factor of 0.9997 at each episode until reaching a minimum threshold of 0.01. During training, the model randomly selects a batch of experiences from memory and trains on that batch. A batch size of 64 was chosen to expedite training while avoiding overfitting or underfitting.

The agent was trained on the synthetic data we prepared. The rewards received by the Deep Reinforcement Learning (DRL) agent were calculated using the following equation:

\[
R_t = \log(\text{reward} + 1) - 1 \quad \text{(16)}
\]

This formulation ensured that the rewards remained bounded and prevented potential convergence issues due to excessively high returns. Additionally, to prevent infinite rewards in cases where the net return during an instance was zero, the reward was truncated by one.

This training configuration aimed to optimize the Q-values for intelligent decision-making in precise food spoilage tracking control within the overall agricultural and transportation environment. Through iterative training episodes and meticulous reward calculation, the DRL agent learned to navigate the environment effectively, facilitating optimal food spoilage tracking strategies for spoilage prevention and management.

\subsection{Metrics for Evaluation}

To evaluate the effectiveness of our Hybrid Deep Q-Learning Framework for precision food spoilage tracking, we employed a set of carefully chosen performance metrics. These metrics comprehensively assess the predictive accuracy, learning efficiency, and training dynamics of the proposed model.

\textbf{Spoilage Accuracy} gives the percentage of correct spoilage level predictions made by the agent compared to the true spoilage levels determined by the environment's logic.

\[
\text{Spoilage Accuracy} = \frac{\text{Total Predictions}}{\text{Correct Predictions}} \times 100 \quad \text{(17)}
\]

It helps evaluate the system's predictive capability in classifying spoilage levels accurately.

\textbf{Reward-to-Step Ratio} gives the average reward received per action step during training. It reflects how well the agent maximizes its cumulative reward over time.

\[
\text{Reward-to-Step Ratio} = \frac{\sum \text{Rewards Collected}}{\text{Number of Steps Taken}} \quad \text{(18)}
\]

It indicates the overall effectiveness of the reinforcement learning agent in optimizing rewards per decision.

\textbf{Loss Decrease Rate} gives the rate of decrease in the loss function over the course of training, showing how effectively the model converges.

\[
\text{Loss Decrease Rate} = \frac{\text{Initial Loss} - \text{Final Loss}}{\text{Number of Steps}} \quad \text{(19)}
\]

This metric helps in assessing the stability and speed of model convergence during training.

\textbf{Exploration Rate Decay} gives out the decay rate of the epsilon parameter in the epsilon-greedy policy, indicating the balance between exploration and exploitation.

\[
\text{Exploration Rate Decay} = \epsilon_{\text{current}} \times \epsilon_{\text{decay}} \quad \text{(20)}
\]

It demonstrates how the agent transitions from exploring actions to exploiting learned policies.

The performance metrics used in our framework provide valuable insights into the model's efficacy and learning dynamics. Spoilage Accuracy reflects the robustness of the trained agent in accurately identifying spoilage levels, aligning with the goals of precision tracking. Reward-to-Step Ratio indicates the agent's effectiveness in learning and making decisions that maximize rewards within the defined structure. Loss Decrease Rate highlights the agent's ability to improve predictions over time, as evidenced by a consistent decline in loss during training. Finally, Exploration Rate Decay ensures a balanced transition from exploration to exploitation, calibrated to avoid premature convergence (insufficient exploration) or prolonged randomness (over-exploration). Together, these metrics provide a comprehensive evaluation of the framework's performance.

\section{Results and Discussion}
The effectiveness and applicability of the proposed Hybrid Deep Q-Learning framework were evaluated using both synthetically generated datasets and real-time sensor data from an IoT-enabled hardware setup. This comprehensive testing strategy was designed to examine not only the predictive accuracy but also the interpretability and interpretability of the system under both controlled and real-world conditions. Synthetic data simulations allowed the model to experience a wide range of spoilage scenarios by varying environmental parameters such as temperature, humidity, gas concentrations, and soil moisture, enabling the agent to learn generalized policies for spoilage tracking. The resulting actions were interpretable, as each was contextually mapped to identifiable environmental triggers and associated spoilage risks. Subsequently, real-time data from Arduino-based hardware sensors validated the framework’s robustness and practical applicability. Sensor feedback involving dynamic changes—such as rising alcohol and methane gas levels, or fluctuating temperature and humidity—demonstrated the model's capacity to respond effectively through clearly defined and interpretable actions, ranging from routine monitoring to emergency alerts. The use of LSTM and RNN components further enriched the interpretability by enabling the model to capture temporal patterns over extended sequences, leading to better decision reasoning and anticipatory responses. This capability was particularly critical in real-time spoilage prediction, where gradual trends such as sustained increases in gas emissions provided early indicators of deterioration. The alignment of reinforcement learning actions with interpretable spoilage levels ensures transparency, while the successful application across both data types underscores the model’s real-time deployability and hardware compatibility. Overall, the proposed framework delivers a scalable and intelligent solution that not only ensures robust spoilage detection but also supports transparent decision-making, contributing to improved food safety and waste reduction in IoT-enabled environments.

\subsection{Experimental Setup }
We performed an empirical assessment of the proposed approaches by integrating reinforcement learning (RL) frameworks and simulations, executed in Python on a Windows 10 setup. The hardware used for the experiments included a Ryzen 5 5000 series CPU with 8GB of RAM. All tests were carried out with Python 3.10 on Google Collaboratory.

\subsection{Dataset Description }
In the Evaluation of the DRL Agent Using Synthetic Data, the synthetic data generation process aimed to closely mimic real-time sensor data, training with 1000 data instances. The following considerations were incorporated into the synthetic data generation:
\begin{itemize}
    \item \textbf{Temperature:} Generated using a normal distribution with a mean of 25°C and a standard deviation of 5°C to emulate typical environmental temperature fluctuations.
    \item \textbf{Humidity:} Simulated using a normal distribution with a mean of 60\% and a standard deviation of 10\%, representing varying levels of atmospheric moisture.
    \item \textbf{Moisture:} Mimicked through a normal distribution with a mean of 200 and a standard deviation of 50, resembling soil moisture content across different scenarios.
    \item \textbf{MQ3 Sensor Data:} Generated using a normal distribution with a mean of 150 and a standard deviation of 30, representing alcohol concentrations in the environment.
    \item \textbf{MQ4 Sensor Data:} Simulated using a normal distribution with a mean of 250 and a standard deviation of 30, indicative of methane gas concentrations.
    \item \textbf{Noise Addition:} 
    \begin{itemize}
        \item Random noise was introduced to each sensor reading to emulate real-world variability.
        \item The noise, sampled from a normal distribution with a mean of 0 and a standard deviation of 5, introduced subtle fluctuations to the sensor data, reflecting inherent environmental uncertainties.
    \end{itemize}
    \item \textbf{Synthetic Dataset:} 
    \begin{itemize}
        \item Structured as a pandas DataFrame to capture the complexity and dynamics of environmental conditions relevant to precise food spoilage tracking control.
        \item Ensured that the DRL agent's training and evaluation leveraged realistic data representations, facilitating robust performance analysis and validation.
    \end{itemize}
    \item \textbf{Spoilage Thresholds for Adverse Conditions:}
    \begin{itemize}
        \item \textbf{Temperature:} 30°C – Elevated temperatures impacting soil moisture dynamics.
        \item \textbf{Humidity:} 70\% – High atmospheric moisture levels affecting transpiration rates.
        \item \textbf{Moisture:} 250 – Excessive soil moisture content hindering root respiration.
        \item \textbf{MQ3 Sensor:} 180 – Elevated alcohol concentrations suggestive of fermentation or decomposition.
        \item \textbf{MQ4 Sensor:} 280 – Heightened methane gas levels indicative of anaerobic conditions or organic matter decomposition.
    \end{itemize}
    \item \textbf{Real-Time Data Collection:}
    \begin{itemize}
        \item Spoilage thresholds derived from empirical observations and literature review:
        \begin{itemize}
            \item \textbf{Temperature:} 28.5°C
            \item \textbf{Humidity:} 92\%
            \item \textbf{MQ3:} 270
            \item \textbf{MQ4:} 340
        \end{itemize}
        \item \textbf{Statistical Analysis of Collected Data:}
        \begin{itemize}
            \item \textbf{Temperature:} Mean = 28.29°C, Standard deviation = 0.19°C.
            \item \textbf{Humidity:} Mean = 91.87\%, Standard deviation = 0.53\%.
            \item \textbf{MQ3:} Mean = 259.89, Standard deviation = 28.97.
            \item \textbf{MQ4:} Mean = 272.54, Standard deviation = 66.86.
        \end{itemize}
        
    \end{itemize}
\end{itemize}
Figure 4 shows the real time setup of the data collection.

\begin{figure}[htbp]
    \centering
    \includegraphics[width=0.8\linewidth]{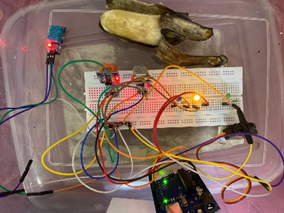} % Change filename and width accordingly
    \caption{ Real time photo of the sensors setup for the data collection}
    \label{fig:your_label}
\end{figure}

\subsection{Evaluation of the DRL Agent Using Synthetic and Real-Time Sensor Data}
We will evaluate the performance of the Deep Reinforcement Learning (DRL) agent using both synthetic data generated from environmental models and real-time sensor data collected from agricultural settings. This evaluation aims to validate the effectiveness of the proposed DRL framework in optimizing food spoilage tracking strategies and improving crop cultivation outcomes. For the simulated data experiments, we will run the agents for 1000 episodes, as this extended duration allows the models to thoroughly explore and learn from the synthetic environment, enabling a more comprehensive understanding of the action-reward dynamics and convergence behavior. In contrast, for real-time hardware data, we will limit the agents to 100 episodes. This reduced duration is justified by the inherent constraints of real-world systems, such as resource availability, computational overhead, and the need for quicker decision-making. Additionally, hardware data often reflects real-time feedback, which accelerates learning and reduces the necessity for extended iterations compared to simulations.

\subsection{Evaluation of the DRL Agent Using Synthetic Data with other agent frameworks}
In this evaluation, we compare the performance of our proposed Deep Reinforcement Learning (DRL) agent with several other agent frameworks: LSTM, RNN, Monte Carlo, and a basic Artificial Neural Network (ANN). Each of these methods offers different strengths depending on the complexity of the task. The LSTM + RNN hybrid model leverages the power of both LSTM’s long-term memory and RNN’s sequential processing, making it ideal for dynamic tasks like spoilage prediction. The basic ANN framework, which consists of dense layers and ReLU activations, provides a simpler approach that can serve as a good baseline for tasks with less complex data dependencies. LSTM, focused on capturing long-term dependencies, excels in time-series prediction but can be more computationally intensive. The RNN, while simpler than LSTM, also handles sequential data but struggles with long-term dependencies, making it less effective in certain dynamic scenarios. Finally, the Monte Carlo method uses random sampling to simulate possible outcomes, providing an alternative approach when uncertainty is high, though it may not perform as well in structured tasks like spoilage prediction. Each framework is evaluated based on key metrics such as spoilage accuracy, reward-to-step ratio, loss decrease rate, and spoilage class distribution. The whole simulation was run for 1000 episodes.
The comparative analysis of the various agent frameworks—LSTM+RNN, ANN, LSTM, RNN, and Monte Carlo—on key metrics (Spoilage Accuracy, Reward-to-Step Ratio, Loss Decrease Rate, Exploration Rate Decay, and Spoilage Class Distribution) provides valuable insights into the performance of each model within the context of food spoilage prediction.
Spoilage Accuracy reflects how accurately each agent predicts the level of food spoilage. The LSTM+RNN model, with a spoilage accuracy of 0.82, performs the best, indicating its ability to capture both temporal dependencies and sequential patterns, which is crucial in predicting spoilage trends. This is expected, as the hybrid LSTM-RNN model leverages LSTM's capability to store long-term information and RNN's processing of sequential data, making it ideal for this dynamic environment. The ANN, at 0.74, shows good accuracy but struggles to capture long-term temporal relationships, hence slightly lower performance. LSTM’s accuracy of 0.77 demonstrates its ability to handle sequential data well but still falls behind the combined LSTM+RNN approach. RNN, with an accuracy of 0.75, exhibits moderate performance, but its inability to model long-term dependencies effectively limits its accuracy. The Monte Carlo method shows an accuracy of 0.00, which suggests that it is not suitable for structured spoilage prediction tasks, as it primarily uses random sampling and does not consider temporal data dependencies.
Reward-to-Step Ratio measures the efficiency of the agent in achieving rewards relative to the number of actions taken. A higher value signifies a more efficient agent. Here, the Monte Carlo method has the highest ratio at 0.7, which can be attributed to its stochastic nature of random sampling and exploration. However, this high ratio does not imply optimal performance since the reward is not necessarily correlated with the actual spoilage prediction. The proposed LSTM+RNN agent comes next with a reward-to-step ratio of 0.63, signifying that it efficiently balances exploration and exploitation in decision-making. The ANN model (0.49) and LSTM (0.54) have lower ratios, reflecting a less efficient exploration-exploitation balance, likely due to limitations in how these networks process the sequential food spoilage data. The RNN has a slightly lower ratio at 0.51, which is expected as it tends to struggle with longer-term dependencies, resulting in less efficient exploration.
Loss Decrease Rate metric indicates how fast the agent reduces its error over time. The LSTM model has the highest loss decrease rate of 0.00143, suggesting that it improves the most rapidly. This is understandable because LSTM is designed to handle time-series data and sequential dependencies, which aligns well with the gradual improvement of spoilage predictions. The RNN follows closely with a rate of 0.00089, showing decent improvement but not as fast as LSTM. The proposed LSTM+RNN model has a negative loss decrease rate of -0.00042, indicating that the loss increases slightly over time. This might be due to overfitting or the model reaching a point of diminishing returns, where the improvement slows down, possibly due to the combined nature of LSTM and RNN components. The ANN, with a much higher loss decrease rate of 0.42156, demonstrates a quick reduction in loss at the beginning but may struggle in long-term performance, as reflected in its lower spoilage accuracy. Monte Carlo does not have a loss decrease rate listed, which further suggests its inapplicability in this task since it relies on random sampling rather than a model-based approach.
Exploration Rate Decay metric is crucial in reinforcement learning as it dictates how exploration (trying new actions) decays over time. A higher decay indicates that the agent gradually focuses more on exploitation (choosing the best-known action). In all cases, the exploration rate decay is identical at 0.995, meaning that the agents are designed to reduce exploration over time and rely more on their learned policy as training progresses. This is consistent across all models, suggesting that they all employ a similar strategy for exploration-exploitation balance as they learn.
The spoilage class distribution provides insights into how the agent classifies different levels of spoilage. For instance, in the proposed LSTM+RNN model, the majority of predictions (817) fall into the '1' class (low spoilage), with fewer predictions for moderate (3), high (0), and severe spoilage (2). This indicates a skew towards less spoilage, likely because the agent learns to predict low spoilage more frequently. The ANN model shows a similar distribution but is slightly more evenly distributed across the classes, suggesting a better balance in spoilage prediction. The LSTM model also tends to classify more frequently into the '1' class, while the RNN shows a somewhat more balanced distribution, although still leaning towards lower spoilage predictions. Monte Carlo, on the other hand, distributes the spoilage across classes more evenly but lacks any predictive power, as its accuracy is 0.00.
Based on the evaluation of all the metrics, the LSTM+RNN agent framework emerges as the best-performing agent for the task of food spoilage tracking. Its high spoilage accuracy (0.82) and efficient reward-to-step ratio (0.63) indicate that it excels in both making accurate spoilage predictions and learning efficiently. While the loss decrease rate for this agent shows a slight increase in error over time, this is outweighed by its superior predictive performance compared to the other models. The LSTM+RNN framework's ability to capture both short-term and long-term dependencies is critical for dynamic tasks like spoilage tracking, making it the most suitable choice for real-time decision-making in food storage environments. Table 2 presents the values for each metric across the various agent frameworks. Figures 5, 6, and 7 display the reward per episode, loss per episode, and spoilage class distribution for the LSTM+RNN agent, respectively. Figures 8, 9, and 10 show the reward per episode, loss per episode, and spoilage class distribution for the ANN agent. Figures 11, 12, and 13 illustrate the reward per episode, loss per episode, and spoilage class distribution for the LSTM agent. Figures 14, 15, and 16 present the reward per episode, loss per episode, and spoilage class distribution for the RNN agent. Finally, Figures 17 and 18 depict the reward per episode and spoilage class distribution for the Monte Carlo method.
\begin{figure}[htbp]
    \centering
    \includegraphics[width=0.8\linewidth]{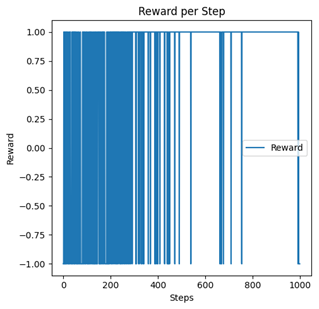} % Change filename and width accordingly
    \caption{ Reward plot for the LSTM+RNN Agent in the simulated data environment. Here, "steps" represent episodes, and the agent runs for a total of 1000 episodes.}
    \label{fig:your_label}
\end{figure}

\begin{figure}[htbp]
    \centering
    \includegraphics[width=0.8\linewidth]{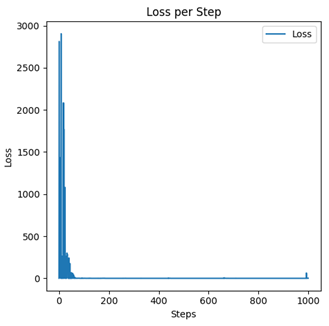} % Change filename and width accordingly
    \caption{Loss plot for the LSTM+RNN Agent in the simulated data environment. Here, "steps" represent episodes, and the agent runs for a total of 1000 episodes.}
    \label{fig:your_label}
\end{figure}

\begin{figure}[htbp]
    \centering
    \includegraphics[width=0.8\linewidth]{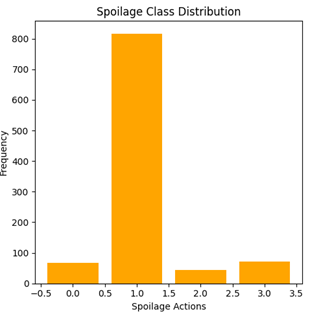} % Change filename and width accordingly
    \caption{Spoilage Class Distribution for the LSTM+RNN Agent in the simulated data environment, for all 4 possible actions}
    \label{fig:your_label}
\end{figure}

\begin{figure}[htbp]
    \centering
    \includegraphics[width=0.8\linewidth]{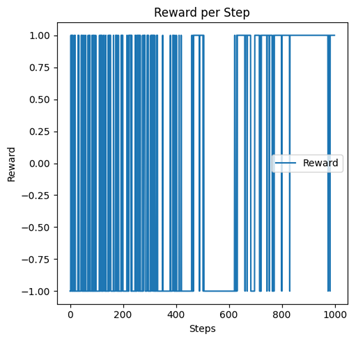} % Change filename and width accordingly
    \caption{Reward plot for the ANN Agent in the simulated data environment. Here, "steps" represent episodes, and the agent runs for a total of 1000 episodes.}
    \label{fig:your_label}
\end{figure}

\begin{figure}[htbp]
    \centering
    \includegraphics[width=0.8\linewidth]{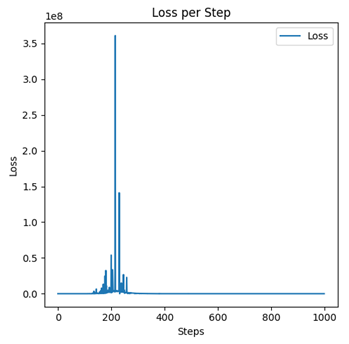} % Change filename and width accordingly
    \caption{Loss plot for the ANN Agent in the simulated data environment. Here, "steps" represent episodes, and the agent runs for a total of 1000 episodes.}
    \label{fig:your_label}
\end{figure}

\begin{figure}[htbp]
    \centering
    \includegraphics[width=0.8\linewidth]{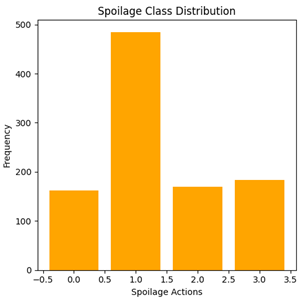} % Change filename and width accordingly
    \caption{Spoilage Class Distribution for the ANN Agent in the simulated data environment, for all 4 possible actions}
    \label{fig:your_label}
\end{figure}

\begin{figure}[htbp]
    \centering
    \includegraphics[width=0.8\linewidth]{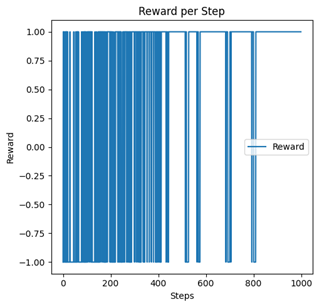} % Change filename and width accordingly
    \caption{ Reward plot for the LSTM Agent in the simulated data environment. Here, "steps" represent episodes, and the agent runs for a total of 1000 episodes.}
    \label{fig:your_label}
\end{figure}

\begin{figure}[htbp]
    \centering
    \includegraphics[width=0.8\linewidth]{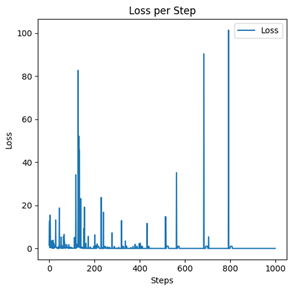} % Change filename and width accordingly
    \caption{Loss plot for the LSTM Agent in the simulated data environment. Here, "steps" represent episodes, and the agent runs for a total of 1000 episodes. }
    \label{fig:your_label}
\end{figure}

\begin{figure}[htbp]
    \centering
    \includegraphics[width=0.8\linewidth]{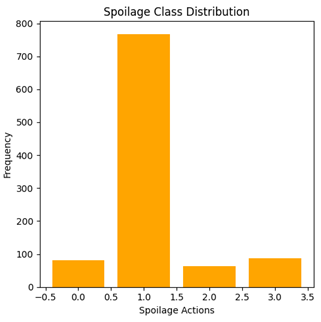} % Change filename and width accordingly
    \caption{ Spoilage Class Distribution for the LSTM Agent in the simulated data environment, for all 4 possible actions}
    \label{fig:your_label}
\end{figure}

\begin{figure}[htbp]
    \centering
    \includegraphics[width=0.8\linewidth]{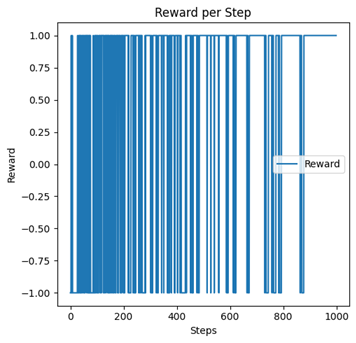} % Change filename and width accordingly
    \caption{Reward plot for the RNN Agent in the simulated data environment. Here, "steps" represent episodes, and the agent runs for a total of 1000 episodes. }
    \label{fig:your_label}
\end{figure}

\begin{figure}[htbp]
    \centering
    \includegraphics[width=0.8\linewidth]{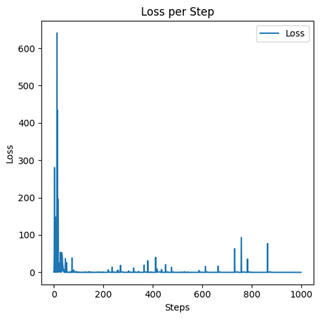} % Change filename and width accordingly
    \caption{Loss plot for the RNN Agent in the simulated data environment. Here, "steps" represent episodes, and the agent runs for a total of 1000 episodes. }
    \label{fig:your_label}
\end{figure}

\begin{figure}[htbp]
    \centering
    \includegraphics[width=0.8\linewidth]{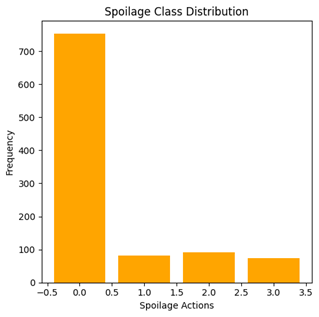} % Change filename and width accordingly
    \caption{Spoilage Class Distribution for the RNN Agent in the simulated data environment, for all 4 possible actions }
    \label{fig:your_label}
\end{figure}

\begin{figure}[htbp]
    \centering
    \includegraphics[width=0.8\linewidth]{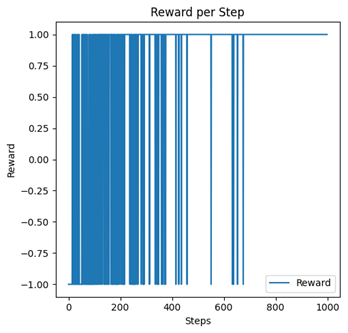} % Change filename and width accordingly
    \caption{Reward plot for the Monte Carlo Agent in the simulated data environment. Here, "steps" represent episodes, and the agent runs for a total of 1000 episodes}
    \label{fig:your_label}
\end{figure}

\begin{figure}[htbp]
    \centering
    \includegraphics[width=0.8\linewidth]{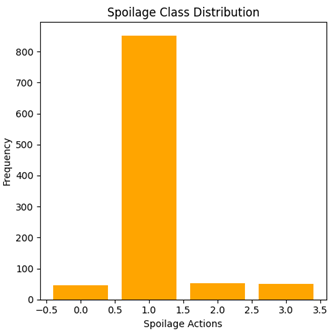} % Change filename and width accordingly
    \caption{Spoilage Class Distribution for the Monte Carlo Agent in the simulated data environment, for all 4 possible actions }
    \label{fig:your_label}
\end{figure}

\subsection{Evaluation of the DRL Agent Using Real-Time Sensor Data with other agent frameworks}
In addition to the synthetic data analysis, we also conducted real-time data validation using the same agent frameworks for comparative analysis. The real-time setup involves sensors that monitor food spoilage conditions, providing continuous data streams for the agents to process and decide the appropriate actions based on real-world inputs. Figure 19 presents the real-time output from the sensor setup, which captures data related to food spoilage conditions such as temperature, humidity, and air quality. Figure 20 shows a graph, generated on the Arduino UI, that interpolates the sensor readings, highlighting the variation and trends of different sensor values over time. These figures complement the previous analysis by providing insight into how each agent performs in a dynamic, real-world environment, further validating the effectiveness of the proposed LSTM+RNN agent framework. Figures 21, 22, and 23 represent the reward per episode, loss per episode, and spoilage action distribution for the proposed LSTM+RNN agent, respectively. Similarly, Figures 24, 25, and 26 depict the reward per episode, loss per episode, and spoilage action distribution for the LSTM-based agent. Figures 27, 28, and 29 show the same metrics for the RNN-based agent. The results for the ANN-based agent are illustrated in Figures 30, 31, and 32, while Figures 33 and 34 present the reward per episode and spoilage action distribution for the Monte Carlo simulation-based agent. Table 3 provides the metrics evaluation for each agent framework tested on real-time hardware data. The only difference from the simulated data evaluation is the exclusion of the spoilage distribution metric. This decision was made because, for hardware data, spoilage distribution served more as informational context rather than a direct performance metric. In contrast, for simulated data, spoilage distribution was included as a metric to enhance clarity and interpretability of the results.

\begin{table*}[htbp]
    \centering
    \caption{Comparative analysis on self-defined metrics for each agent framework}
    \label{tab:comparative_analysis}
    \renewcommand{\arraystretch}{1.5} % Increases row spacing
    \resizebox{0.98\textwidth}{!}{ % Slightly larger table while maintaining page fit
    \begin{tabular}{l c c c c c}
        \toprule
        \textbf{Agent Framework} & \textbf{Spoilage Accuracy} & \textbf{Reward-to-Step Ratio} & \textbf{Loss Decrease Rate} & \textbf{Exploration Rate Decay} & \textbf{Spoilage Class Distribution} \\
        \midrule
        LSTM+RNN (Proposed) & 0.82 & 0.63 & -0.00042 & 0.995 & \{0: 67, 3: 71, 1: 817, 2: 45\} \\
        ANN & 0.74 & 0.49 & 0.42156 & 0.995 & \{1: 745, 2: 98, 0: 83, 3: 74\} \\
        LSTM & 0.77 & 0.54 & 0.00143 & 0.995 & \{0: 81, 1: 768, 3: 87, 2: 64\} \\
        RNN & 0.75 & 0.51 & 0.00089 & 0.995 & \{3: 73, 2: 92, 0: 754, 1: 81\} \\
        Monte Carlo & 0.00 & 0.70 & N/A & 0.995 & \{0: 45, 2: 52, 3: 51, 1: 852\} \\
        \bottomrule
    \end{tabular}
    } % End of resizebox
\end{table*}
\begin{figure}[htbp]
    \centering
    \includegraphics[width=0.8\linewidth]{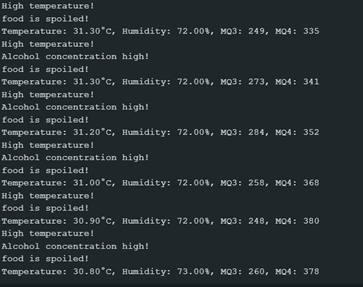} % Change filename and width accordingly
    \caption{ Real time updating and collection of data}
    \label{fig:your_label}
\end{figure}

\begin{figure}[htbp]
    \centering
    \includegraphics[width=0.8\linewidth]{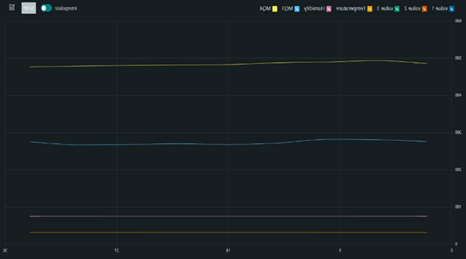} % Change filename and width accordingly
    \caption{Sensor readings on graph}
    \label{fig:your_label}
\end{figure}

\begin{figure}[htbp]
    \centering
    \includegraphics[width=0.8\linewidth]{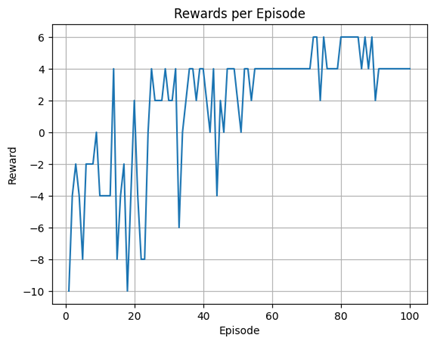} % Change filename and width accordingly
    \caption{Reward plot for the LSTM+RNN Agent in the realtime hardware data environment, running for 100 episodes}
    \label{fig:your_label}
\end{figure}

\begin{figure}[htbp]
    \centering
    \includegraphics[width=0.8\linewidth]{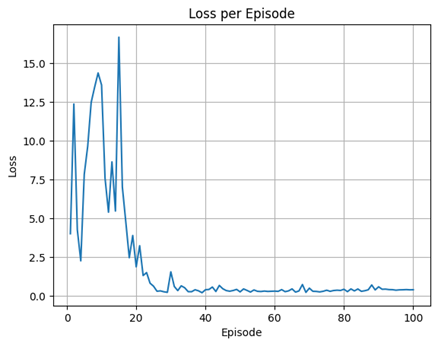} % Change filename and width accordingly
    \caption{Loss plot for the LSTM+RNN Agent in the realtime hardware data environment, running for 100 episodes }
    \label{fig:your_label}
\end{figure}

\begin{figure}[htbp]
    \centering
    \includegraphics[width=0.8\linewidth]{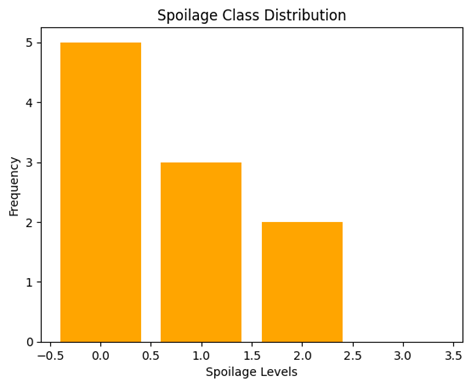} % Change filename and width accordingly
    \caption{Spoilage Class Distribution for the LSTM+RNN Agent in the realtime hardware data environment, running for 100 episodes }
    \label{fig:your_label}
\end{figure}

\begin{figure}[htbp]
    \centering
    \includegraphics[width=0.8\linewidth]{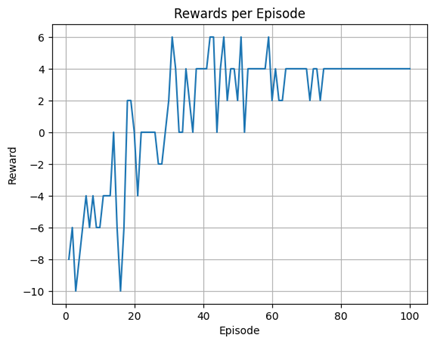} % Change filename and width accordingly
    \caption{Reward plot for the LSTM Agent in the realtime hardware data environment, running for 100 episodes}
    \label{fig:your_label}
\end{figure}

\begin{figure}[htbp]
    \centering
    \includegraphics[width=0.8\linewidth]{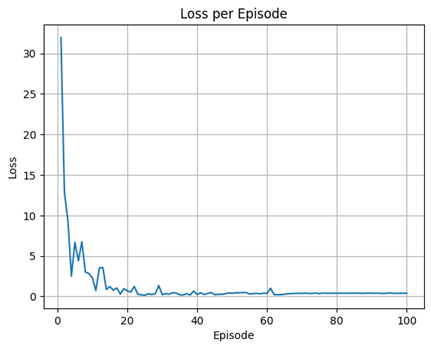} % Change filename and width accordingly
    \caption{Loss plot for the LSTM Agent in the realtime hardware data environment, running for 100 episodes}
    \label{fig:your_label}
\end{figure}

\begin{figure}[htbp]
    \centering
    \includegraphics[width=0.8\linewidth]{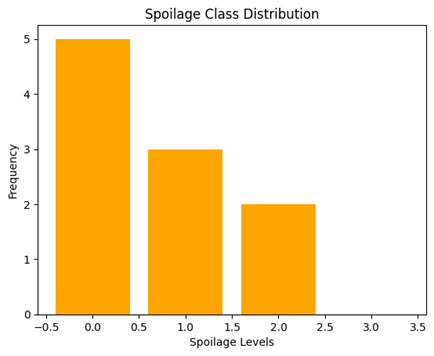} % Change filename and width accordingly
    \caption{: Spoilage Class Distribution for the LSTM Agent in the realtime hardware data environment, running for 100 episodes }
    \label{fig:your_label}
\end{figure}

\begin{figure}[htbp]
    \centering
    \includegraphics[width=0.8\linewidth]{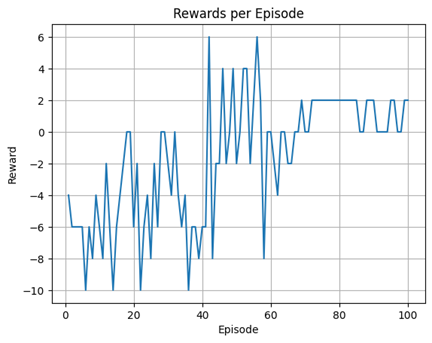} % Change filename and width accordingly
    \caption{Reward plot for the RNN Agent in the realtime hardware data environment, running for 100 episodes}
    \label{fig:your_label}
\end{figure}

\begin{figure}[htbp]
    \centering
    \includegraphics[width=0.8\linewidth]{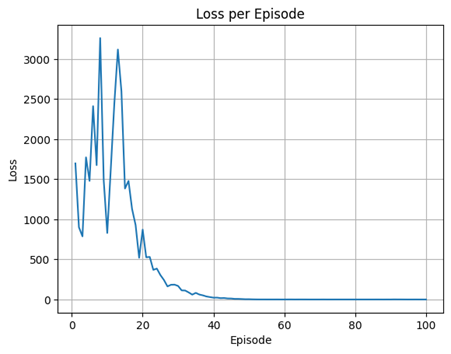} % Change filename and width accordingly
    \caption{: Loss plot for the RNN Agent in the realtime hardware data environment, running for 100 episodes
              }
    \label{fig:your_label}
\end{figure}

\begin{figure}[htbp]
    \centering
    \includegraphics[width=0.8\linewidth]{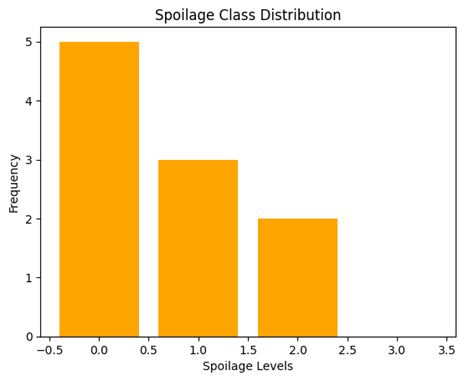} % Change filename and width accordingly
    \caption{Spoilage Class Distribution for the RNN Agent in the realtime hardware data environment, running for 100 episodes }
    \label{fig:your_label}
\end{figure}

\begin{figure}[htbp]
    \centering
    \includegraphics[width=0.8\linewidth]{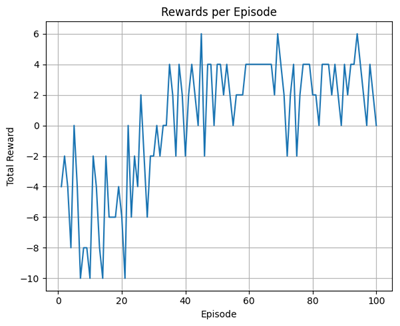} % Change filename and width accordingly
    \caption{ Reward plot for the ANN Agent in the realtime hardware data environment, running for 100 episodes}
    \label{fig:your_label}
\end{figure}

\begin{figure}[htbp]
    \centering
    \includegraphics[width=0.8\linewidth]{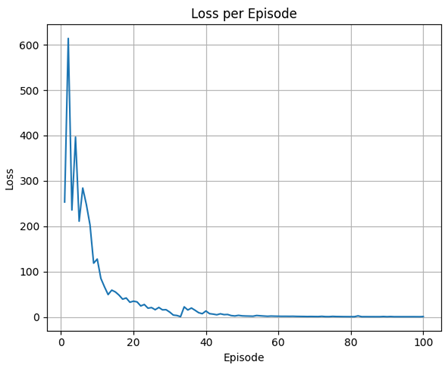} % Change filename and width accordingly
    \caption{Loss plot for the ANN Agent in the realtime hardware data environment, running for 100 episodes}
    \label{fig:your_label}
\end{figure}

\begin{figure}[htbp]
    \centering
    \includegraphics[width=0.8\linewidth]{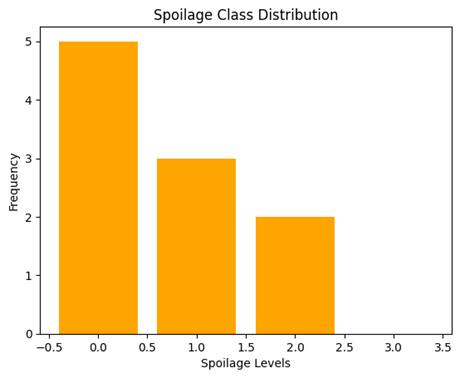} % Change filename and width accordingly
    \caption{Spoilage Class Distribution for the ANN Agent in the realtime hardware data environment, running for 100 episodes}
    \label{fig:your_label}
\end{figure}

\begin{figure}[htbp]
    \centering
    \includegraphics[width=0.8\linewidth]{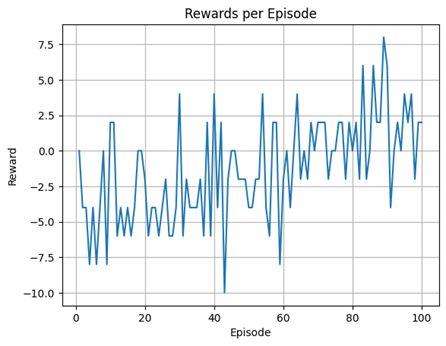} % Change filename and width accordingly
    \caption{Reward plot for the Monte Carlo Agent in the realtime hardware data environment, running for 100 episodes }
    \label{fig:your_label}
\end{figure}

\begin{figure}[htbp]
    \centering
    \includegraphics[width=0.8\linewidth]{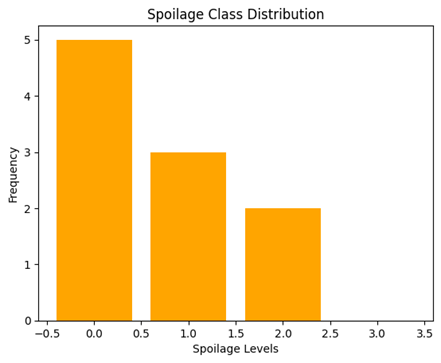} % Change filename and width accordingly
    \caption{Spoilage Class Distribution for the Monte Carlo Agent in the realtime hardware data environment, running for 100 episodes}
    \label{fig:your_label}
\end{figure}

The results in Table 3 provide a comparative evaluation of different agent frameworks on real-time hardware data across four critical metrics: spoilage accuracy, reward-to-step ratio, loss decrease rate, and exploration rate decay. Each metric offers a unique perspective on the performance and reliability of the frameworks in predicting spoilage and optimizing decision-making.
Starting with spoilage accuracy, the combined LSTM+RNN framework achieves the highest value of 0.59, closely followed by LSTM at 0.58. This indicates the ability of these models to accurately predict spoilage levels, which can be attributed to their temporal data processing strengths. In contrast, ANN, while reasonably effective with a spoilage accuracy of 0.55, falls slightly behind due to its inability to fully leverage sequential dependencies in the data. The RNN framework shows a significantly lower spoilage accuracy of 0.42, likely reflecting its limitations in handling long-term dependencies compared to LSTM. Monte Carlo performs the worst in this regard, with a spoilage accuracy of 0.41, emphasizing the limitations of policy-based methods when applied to highly variable real-time data.
The reward-to-step ratio further emphasizes the utility of each framework. LSTM+RNN demonstrates the best reward-to-step ratio at 0.17, reflecting its balanced ability to optimize rewards while minimizing penalized actions. LSTM, with a ratio of 0.16, exhibits similar performance, underscoring the power of LSTM in handling temporal patterns. ANN shows a moderate ratio of 0.11, reflecting its satisfactory yet less efficient ability to align actions with rewards. The negative ratios of RNN (-0.17) and Monte Carlo (-0.18) suggest frequent penalized actions, indicating challenges in aligning predictions with rewards, particularly for dynamic environments like real-time hardware data.
The loss decrease rate provides insight into the learning stability of each framework. LSTM+RNN outperforms with a steady rate of 0.09148, highlighting its capability for consistent learning convergence. LSTM shows a slightly higher rate at 0.12168, indicating quicker learning but potentially higher variance in optimization. ANN presents a negative loss decrease rate of -6.27726, suggesting instability in learning, possibly due to the absence of sequential context modeling. RNN displays a disproportionately large value of 26.60358, reflecting erratic learning dynamics that could hinder reliable convergence. Monte Carlo lacks a measurable loss decrease rate due to its non-differentiable, policy-based nature, limiting direct optimization comparisons.

\begin{table*}[htbp]
    \centering
    \caption{Comparative Analysis on self-defined metrics for each agent framework on hardware real-time data}
    \label{tab:comparative_analysis_hardware}
    \renewcommand{\arraystretch}{1.5} % Increases row spacing for better readability
    \resizebox{0.8\textwidth}{!}{ % Adjusts size to keep the text readable while fitting within the page
    \begin{tabular}{l c c c c}
        \toprule
        \textbf{Agent Framework} & \textbf{Spoilage Accuracy} & \textbf{Reward-to-Step Ratio} & \textbf{Loss Decrease Rate} & \textbf{Exploration Rate Decay} \\
        \midrule
        LSTM + RNN (Proposed) & 0.59 & 0.17 & 0.09148 & 0.80659 \\
        LSTM & 0.58 & 0.16 & 0.12168 & 0.80659 \\
        RNN & 0.42 & -0.17 & 26.60358 & 0.80659 \\
        ANN & 0.55 & 0.11 & -6.27726 & 0.80659 \\
        Monte Carlo & 0.41 & -0.18 & N/A & 0.80659 \\
        \bottomrule
    \end{tabular}
    } % End of resizebox
\end{table*}

Exploration rate decay remains consistent across frameworks, with a uniform value of 0.80659. This uniformity implies that all frameworks employed a comparable exploration-exploitation trade-off strategy during training, allowing for a controlled comparison across other metrics without exploration differences influencing the outcomes.
In analyzing the overall trends, it becomes clear that LSTM+RNN consistently outperforms other frameworks across spoilage accuracy, reward-to-step ratio, and loss decrease rate. Its ability to combine the strengths of LSTM (handling long-term dependencies) and RNN (processing sequential patterns) makes it the most effective agent for hardware data. The closeness of values between LSTM+RNN and LSTM alone suggests that the integration of RNN slightly enhances overall performance, but LSTM’s foundational capabilities play a dominant role. ANN, while decent, lacks the sequential depth offered by LSTM and RNN, leading to less robust performance. RNN and Monte Carlo, on the other hand, show significant shortcomings in handling the complexities of real-time hardware data.
In conclusion, the LSTM+RNN framework demonstrates superior performance, combining accuracy, reward optimization, and stability in learning. Its closeness to LSTM in metrics underscores LSTM's importance, but the addition of RNN provides an edge in capturing more nuanced temporal patterns. This analysis highlights the value of leveraging hybrid sequential models for real-time applications, particularly in environments characterized by dynamic and complex data.

The figure 20 depicts a line graph illustrating temperature and humidity readings over time, potentially including sensor data from MQ-3 and MQ-4 gas sensors. Observations reveal that banana spoilage accelerates significantly when temperatures rise above 13°C (55°F), with temperature readings between 100 and 600 (assuming the scale is in degrees Celsius) suggesting unsuitable storage conditions. Moreover, humidity readings between 20\% and 36\% indicate insufficient humidity, potentially leading to faster dehydration and shriveling of bananas. If included, rising MQ-3 readings might indicate increased ethanol emission as bananas ripen, while increasing MQ-4 readings could suggest elevated methane levels associated with overripe bananas. 
The spoilage distribution for the hardware data does not include any actions corresponding to a spoilage level of 3, as the system's conditions did not meet the threshold for that level. This highlights that the spoilage action distribution serves as a useful tool in interpreting the agent's decision-making process. By examining the distribution, we can better understand the types of actions the agent is taking and their frequency. This analysis provides insights into how the agent is responding to the observed environmental conditions, and how often it selects each spoilage level. In reinforcement learning, such distributions not only reveal the agent's behavior but also offer valuable information for fine-tuning the policy, ensuring that the agent learns optimal actions while considering the real-world constraints of the system.

\section{Conclusion}
This research presents a hybrid Deep Q-Learning framework that successfully integrates reinforcement learning with real-time IoT sensor systems and synthetic data generation to address the challenge of food spoilage prediction and control. The proposed approach leverages calibrated environmental sensors—such as DHT11, MQ3, MQ4, and soil moisture sensors—embedded in an Arduino-based setup to monitor spoilage indicators dynamically. Through a dual-phase validation process using both synthetically generated and real-time data, the framework has demonstrated high reliability, interpretability, and adaptability in capturing spoilage-related trends. By incorporating Long Short-Term Memory (LSTM) and Recurrent Neural Network (RNN) architectures within the Deep Q-Network (DQN), the model effectively captures complex temporal dependencies and delivers interpretable actions aligned with sensor conditions. This interpretability is reinforced through the rule-based classification embedded in the environment, which maps specific sensor thresholds to clearly defined spoilage states, allowing the agent to make context-aware decisions. Performance evaluations—based on spoilage accuracy, reward-to-step ratio, loss minimization, and exploration decay—confirmed the robustness of the agent, particularly the LSTM+RNN configuration, which achieved the highest spoilage accuracy of 0.59. These results underscore the model’s capability to generalize across data domains and its applicability in real-time food spoilage monitoring. The seamless integration of synthetic simulations and hardware deployment not only validates the model’s predictive strength but also enhances its practical deployment potential. Ultimately, this framework lays a strong foundation for building intelligent, transparent, and real-time spoilage prediction systems that can significantly reduce food waste and improve operational decision-making across the food supply chain.

\section{Future Directions and Limitations}
Expanding on this unique and collaborative method to enhance the decision-making process in precision food spoilage tracking, there are many exciting possibilities to explore in the future:
\subsection{Hybrid Reinforcement Learning Architectures}
Hybrid reinforcement learning architectures blend deep reinforcement learning with other machine learning paradigms like model-based reinforcement learning or imitation learning. In this approach, model-based techniques could augment the deep reinforcement learning model by providing a learned or handcrafted environmental model, allowing for more efficient planning and exploration[42]. This combination can improve sample efficiency, as the model-based component can simulate potential future states and outcomes, guiding the exploration process of the deep reinforcement learning agent. Furthermore, imitation learning can be integrated to leverage expert demonstrations or historical data, providing additional guidance during training and potentially accelerating learning in complex environments. This hybrid approach offers the potential to enhance the generalization and robustness of the food spoilage tracking control system by leveraging the complementary strengths of different learning paradigms. Specifically, it could improve the system's ability to adapt to diverse environmental conditions and crop dynamics, leading to more efficient spoilage management and optimized detection.
\subsection{Adaptive Sampling and Active Learning}
Adaptive sampling and active learning techniques involve dynamically adjusting the data collection process to prioritize gathering informative data points crucial for training the reinforcement learning model. By integrating these methods into the data generation process, computational resources can be efficiently allocated to collect data that maximizes the learning progress. This is achieved by continuously assessing the model's uncertainty or exploration-exploitation trade-offs and adapting the sampling strategy accordingly. For instance, the system could focus on exploring regions of the state space where the model's predictions are uncertain or where there is potential for significant learning progress. This approach could significantly accelerate the learning process by reducing the amount of redundant or less informative data collected during training. For example, He et al [43] proposed an adaptive hybrid active learning method for ordinal classification by considering the ordering information In the context of our food spoilage tracking control system, adaptive sampling and active learning could lead to more efficient data collection, enabling the reinforcement learning model to better understand the complex dynamics of the agricultural environment and make more informed food spoilage tracking decisions, ultimately improving system performance and crop yield.
\subsection{Transfer Learning and Domain Adaptation}
Transfer learning and domain adaptation involve leveraging knowledge gained from one task or domain to improve performance on a related task or domain. In the context of our food spoilage tracking control system, this approach entails pre-training the reinforcement learning model on data from a source agricultural environment or crop type and fine-tuning it on data from a target environment or crop. By transferring knowledge learned from the source domain to the target domain, the model can benefit from insights and patterns that generalize across different environments or crops, reducing the need for extensive data collection and training in each specific scenario. Crabtree et al provides a system for transfer learning and domain adaptation using distributable data models [44]. This approach can significantly accelerate the learning process, particularly in cases where data collection in the target domain is limited or costly. Furthermore, transfer learning and domain adaptation can enhance the adaptability and robustness of the food spoilage tracking control system, enabling it to effectively manage food spoilage tracking across diverse agricultural settings and crop types with minimal manual intervention.
\subsection{Model-based Reinforcement Learning}
Model-based reinforcement learning involves learning a predictive model of the environment dynamics, which is then used to simulate future states and rewards. By incorporating explicit environmental models into the reinforcement learning framework, these methods can potentially improve sample efficiency by reducing the need for extensive trial-and-error exploration [45]. This predictive model enables the agent to plan ahead over longer time horizons, anticipating the consequences of its actions and making more informed decisions. Moreover, the use of explicit environmental models enhances the interpretability of the model's decision-making process, as the agent's behavior can be traced back to its predictions about the future states of the environment. In the context of our food spoilage tracking control system, model-based reinforcement learning could lead to more efficient spoilage management strategies by enabling the agent to simulate the effects of different food spoilage tracking actions and anticipate their impact on crop growth and soil moisture levels, ultimately improving system performance and resource utilization.
\subsection{Hierarchical Reinforcement Learning}
Hierarchical reinforcement learning involves decomposing the food spoilage tracking control problem into multiple levels of abstraction or decision-making hierarchies. In this approach, the agent learns high-level strategies for long-term planning and coordination, as well as low-level policies for fine-grained control and adaptation to local conditions [46]. This hierarchical architecture enables the agent to efficiently navigate complex decision spaces by delegating decision-making responsibilities across different levels of abstraction. For instance, at the higher level, the agent may learn overarching food spoilage tracking schedules or crop management strategies, while at the lower level, it may learn precise food spoilage tracking actions based on local sensor readings and environmental conditions. This hierarchical organization allows for more efficient learning and decision-making, as the agent can focus on learning high-level strategies that generalize across different scenarios while also refining its behavior at lower levels to adapt to specific environmental contexts. In the context of our food spoilage tracking control system, hierarchical reinforcement learning could lead to more robust and adaptive food spoilage tracking strategies by enabling the agent to effectively balance long-term goals, such as maximizing crop yield, with short-term considerations, such as responding to sudden changes in weather conditions or soil moisture levels.
\subsection{Multi-objective Reinforcement Learning}
Multi-objective reinforcement learning (MORL) extends the reinforcement learning framework to handle optimization problems with multiple conflicting objectives, such as maximizing crop yield, minimizing spoilage, and reducing energy consumption. MORL algorithms aim to find Pareto-optimal solutions, which represent trade-offs between different objectives, allowing farmers to make informed decisions based on their preferences For instance, Nowe et al [47] propose a novel temporal dierence learning algorithm that integrates the Pareto dominance relation into a reinforcement learning approach. These algorithms typically involve the optimization of a vector-valued objective function, where each objective corresponds to a different aspect of the system's performance. By considering multiple objectives simultaneously, MORL enables the exploration of diverse food spoilage tracking strategies that balance competing goals, such as maximizing productivity while minimizing resource consumption and environmental impact. In our food spoilage tracking control system, MORL could serve as an upgrade by providing a principled framework for decision-making that accounts for the complex trade-offs inherent in agricultural management. By optimizing multiple objectives concurrently, MORL could help farmers achieve more sustainable and resilient food spoilage tracking practices tailored to their specific needs and priorities.
\subsection{Online Learning and Adaptive Control}
Online learning and adaptive control involve the development of algorithms that enable the reinforcement learning model to continuously adapt to changing environmental conditions and crop dynamics in real-time. These algorithms facilitate dynamic adjustment of food spoilage tracking decisions based on incoming sensor data and feedback from the farming system. Techniques for online policy optimization, reinforcement learning with expert feedback, and adaptive control mechanisms are explored to enhance the model's responsiveness and adaptability. By leveraging real-time data and feedback, the model can dynamically update its policies to optimize food spoilage tracking strategies, ensuring efficient spoilage management and crop growth even in fluctuating conditions. This approach represents an upgrade to our current work by enabling the food spoilage tracking control system to operate in a more autonomous and adaptive manner, continually improving its performance and responsiveness to changing environmental factors[48]. In our system, online learning and adaptive control mechanisms play a crucial role in maintaining optimal food spoilage tracking practices and maximizing crop yield while minimizing resource usage and environmental impact.
\subsection{Robustness and Uncertainty Quantification}Robustness and uncertainty quantification methods focus on enhancing the reliability and safety of the reinforcement learning framework by addressing uncertainties in sensor measurements, environmental dynamics, and model predictions. Techniques such as uncertainty estimation, robust optimization, and adversarial training are incorporated to quantify and mitigate uncertainties in the system[49]. Uncertainty estimation methods assess the confidence of model predictions and sensor measurements, allowing the system to make more informed decisions under uncertainty. Robust optimization techniques optimize food spoilage tracking strategies while considering uncertainties, ensuring performance across diverse operating conditions. Adversarial training enhances the model's resilience to adversarial perturbations, safeguarding against potential threats or unexpected disturbances. By incorporating these methods, our system can improve its adaptability and reliability, ensuring optimal food spoilage tracking control even in uncertain or challenging environments. This upgrade enhances the overall performance and resilience of our food spoilage tracking control system, contributing to more efficient spoilage management and crop cultivation while minimizing risks associated with uncertainties.
While our hybrid Deep Q-Learning framework exhibits promising results, some limitations must be considered. Despite efforts to replicate real-world sensor data through synthetic data generation, there may be scenarios or environmental conditions not fully represented in the dataset, potentially affecting the model's ability to generalize to unseen data. To address this, ongoing research should focus on improving the diversity and representativeness of the synthetic dataset, incorporating a broader range of environmental conditions and scenarios. Additionally, techniques such as transfer learning could be explored to enhance the model's ability to generalize to new and unseen data. Despite calibration protocols, sensor accuracy may degrade over time, leading to inaccurate readings and potentially compromising the effectiveness of our food spoilage tracking system. Continuous monitoring and periodic recalibration of sensors are essential to maintain accuracy and reliability. Additionally, exploring advanced sensor technologies with improved accuracy and longevity could mitigate this limitation. The framework's performance heavily relies on the accuracy and reliability of sensor data. Environmental changes, such as extreme weather conditions or sensor malfunctions, could significantly impact the system's effectiveness. Implementing robust error detection and correction mechanisms within the framework could help identify and mitigate the effects of environmental changes and sensor malfunctions in real-time. The hybrid Deep Q-Learning framework employs complex algorithms and neural network architectures, offering significant advantages but increasing computational complexity and resource requirements. While these advanced techniques provide enhanced predictive capabilities, they also demand substantial computational resources for training and inference. Exploring optimization techniques and model compression methods could help reduce the computational burden without compromising performance. The framework's effectiveness is highly dependent on the quality of the training data. Inaccuracies or biases in the synthetic dataset could lead to suboptimal performance and decision-making by the model. Enhancing data quality through rigorous validation processes, including outlier detection and error correction, is crucial. Additionally, incorporating real-world data into the training process could help improve the model's performance and generalizability. Addressing these limitations necessitates further research and development, including enhancements in data generation techniques, sensor technologies, and model robustness. Additionally, continuous monitoring and maintenance of the system are essential to ensure its long-term effectiveness and reliability in real-world agricultural settings.



\maketitle

\bibliographystyle{plain}   % Use a standard style (e.g., plain, IEEE, or your preferred style)

\end{document}